\documentclass[preprint,12pt,sort&compress]{elsarticle}

\usepackage{booktabs} 
\usepackage{amssymb}
\usepackage{url}
\usepackage{lineno}
\usepackage{hyperref}
\usepackage{epsfig}
\usepackage{graphicx}
\usepackage{subfig}
\usepackage{amsmath}
\usepackage{amssymb}
\usepackage{multirow}
\usepackage{float}
\usepackage{cutwin}
\usepackage{algorithm}
\usepackage{algorithmicx}
\usepackage{algpseudocode}
\usepackage{makecell}
\usepackage{xcolor}
\usepackage{colortbl}
\usepackage{xspace}

\definecolor{Gray}{gray}{0.85}
\newcolumntype{a}{>{\columncolor{Gray}}r}

\newcommand{\sys}{VLQ-ADC\xspace}
\journal{Future Generation Computer Systems}

\begin{document}

\begin{frontmatter}
\title{Vector and Line Quantization for Billion-scale Similarity Search on GPUs}

\author[mymainaddress]{Wei Chen}

\author[mymainaddress,mysecondaryaddress]{Jincai Chen\corref{cor1}}
\ead{jcchen@hust.edu.cn}

\author[address3]{Fuhao Zou\corref{cor1}}

\cortext[cor1]{Corresponding author}
\ead{fuhao\_zou@hust.edu.cn}

\author[monash]{Yuan-Fang Li}

\author[mymainaddress,mysecondaryaddress]{Ping Lu}

\author[mymainaddress]{Qiang Wang}

\author[mysecondaryaddress]{Wei Zhao}

\address[mymainaddress]{Wuhan National Laboratory for Optoelectronics, Huazhong University of Science and Technology, Wuhan 430074, China}

\address[mysecondaryaddress]{Key Laboratory of Information Storage System of Ministry of Education, School of Computer Science and Technology, Huazhong University of Science and Technology, Wuhan 430074, China
}

\address[address3]{School of Computer Science and Technology, Huazhong University of Science and Technology, Wuhan 430074,China}

\address[monash]{Faculty of Information Technology, Monash University, Clayton 3800, Australia}

\begin{abstract}
Billion-scale high-dimensional approximate nearest neighbour (ANN) search has become an important problem for searching similar objects among the vast amount of images and videos available online. The existing ANN methods are usually characterized by their specific indexing structures, including the inverted index and the inverted multi-index structure. The inverted index structure is amenable to GPU-based implementations, and the state-of-the-art systems such as Faiss are able to exploit the massive parallelism offered by GPUs. However, the inverted index requires high memory overhead to index the dataset effectively. The inverted multi-index structure is difficult to implement for GPUs, and also ineffective in dealing with database with different data distributions. In this paper we propose a novel hierarchical inverted index structure generated by vector and line quantization methods. Our quantization method improves both search efficiency and accuracy, while maintaining comparable memory consumption. This is achieved by reducing search space and increasing the number of indexed regions.

We introduce a new ANN search system, VLQ-ADC, that is based on the proposed inverted index, and perform extensive evaluation on two public billion-scale benchmark datasets SIFT1B and DEEP1B. Our evaluation shows that VLQ-ADC significantly outperforms the state-of-the-art GPU- and CPU-based systems in terms of both accuracy and search speed. The source code of VLQ-ADC is available at \url{https://github.com/zjuchenwei/vector-line-quantization}.
\end{abstract}

\begin{keyword}
Quantization; Billion-scale similarity search; high dimensional data; Inverted index; GPU
\end{keyword}

\end{frontmatter}



\section{Introduction}
\label{s1}
In the age of the Internet, the amount of images and videos available online increases incredibly fast and has grown to an unprecedented scale. Google processes over 40,000 various queries per second, and handles more than 400 hours of YouTube video uploads every minute~\cite{Chan2018Fully}. Every day, more than 100 million photos/videos are uploaded to Instagram, more than 300 million uploaded to Facebook, and a total of 50 billion photos have been shared to Instagram\footnote{\url{https://www.omnicoreagency.com/instagram-statistics/}}. As a result, scalable and efficient search for similar images and videos on the billion scale has become an important problem and it has been under intense investigation. 

As online images and videos are unstructured and usually unlabeled, it is hard to compare them directly. A feasible solution is to use real-valued, high-dimensional vectors to represent images and videos, and compare the distances between the vectors to find the nearest ones. Due to the curse of dimensionality~\cite{weber1998quantitative}, it is impractical for multimedia applications to perform exhaustive search in billion-scale datasets. Thus, as an alternative, many \emph{approximate nearest neighbor} (ANN) search algorithms are now employed to tackle the billion-scale search problem for high-dimensional data.
Recent best-performing billion-scale retrieval systems~\cite{Babenko2014Improving,wieschollek2016efficient,Yandex2016Efficient,johnson2017billion,Baranchuk2018Revisiting,yu2015performance} typically utilize two main processes: indexing and encoding.

To avoid expensive exhaustive search, these systems use \emph{index structures} that can partition the dataset space into a large number of disjoint regions, and the search process only collects points from the regions that are closest to the query point. The collected points then form a short list of candidates for each query point. The retrieval system then calculates the distance between each candidate and the query point, and sort them accordingly.

To guarantee query speed, the indexed points need to be loaded into RAM. For large datasets that do not fit in RAM, dataset points are \emph{encoded} into a compressed representation. Encoding has also proven to be critical for memory-limited devices such as GPUs that excel at handling data-parallel tasks. A high-performance CPU like Intel Xeon Platinum 8180 (2.5 GHz, 28 cores) performs 1.12 TFLOP/s single precision peak
performance\footnote{\url{https://ark.intel.com/content/www/us/en/ark/products/120496/intel-xeon-platinum-8180-processor-38-5m-cache-2-50-ghz.html}}. In contrast, GPUs like NVdia Tesla P100 can provide up to 10T FLOP/s single precision peak
performance\footnote{\url{https://images.nvidia.com/content/tesla/pdf/nvidia-tesla-p100-PCIe-datasheet.pdf}}, and are good choices for high performance similarity search systems. Many encoding methods have been proposed, including hashing methods and quantization methods. Hashing methods encode data points to compact binary codes through a hash function \cite{Gong2013Iterative,He2015Deep}, and quantization methods, typically product quantization (PQ), map data points to a set of centroids and use the indices of the centroids to encode the data points~\cite{J2011Product,Kalantidis2014Locally}. By hashing methods, the distance between two data points can be approximated by the Hamming distance between their binary code. By quantization methods, the Euclidean distance between the query and compressed points can be computed efficiently. It has been shown in the literature that quantization encoding can be more accurate than various hashing methods \cite{J2011Product,muja2014scalable,Norouzi2013Cartesian}.

\citet{J2011Product} first introduced an index structure that is able to handle billion-scale datasets effieciently. It is based on the inverted index structure that partitions the high dimensional vector space into Voronoi regions for a set of centroids obtained by a quantization method called vector quantization (VQ) \cite{linde1980algorithm}. This system, called IVFADC, achieves reasonable recall rates in several tens of milliseconds. However, the VQ-based index structure needs to store a large set of full dimensional centroids to produce a huge number of regions, which would require a large amount of memory.

An improved inverted index structure called the inverted multi-index (IMI) was later proposed by \citet{babenko2012inverted}. The IMI is based on product quantization (PQ), which divides the point space into several orthogonal subspaces and clusters the subspaces into Voronoi regions independently. The Cartesian product of regions in each subspace forms regions in the global point space. The strength of the IMI is that it can produce a huge number of regions with much smaller codebooks than that of the inverted index. Due to the huge number of indexed regions, the point space is finely partitioned and each regions contains fewer points. Hence the IMI can provide accurate and concise candidate lists with memory and runtime efficiency.

However, it has been observed that for some billion-scale datasets, the majority of the IMI regions contain no points~\cite{Yandex2016Efficient}, which is a waste of index space and has a negative impact on the final retrieval performance. The reasons for this deficiency is that the IMI learns the centroids independently on the subspaces which are not statistically independent~\cite{Baranchuk2018Revisiting}. In fact, some convolutional neural networks (CNN) produce feature vectors with considerable correlations between the subspaces~\cite{He2015Deep,Razavian2014CNN,Gong2014Multi}.

The high level of parallelism provided by GPUs has recently been leveraged to accelerate similarity search of high-dimensional data, and it has been demonstrated that GPU-based systems are more efficient than CPU-based systems by a large margin~\cite{johnson2017billion,wieschollek2016efficient}. Comparing to IMI structure, the inverted indexing structure proposed by \citet{J2011Product} is more straightforward to parallelize, because the IMI structure depends on a complicated multi-sequence
algorithm, which is sequential in nature \cite{wieschollek2016efficient} and hard to parallelize.

To the best of our knowledge, there are two high performance systems that are able to handle ANN search for billion-scale datasets on the GPU: PQT~\cite{wieschollek2016efficient} and Faiss~\cite{johnson2017billion}. PQT proposes a novel quantization method call line quantization (LQ) and is the first billion-scale similarity retrieval system on the GPU. Subsequently Faiss implements the idea of IVFADC on GPUs and currently has the state-of-the-art performance on GPUs. We compare our method against Faiss and two other systems in Section \ref{sec:s5}.

In this paper, we present VLQ-ADC, a novel billion-scale ANN similarity search framework. VLQ-ADC includes a two-level hierarchical inverted indexing structure based on Vector and Line Quantization (VLQ), which can be implemented on GPU efficiently. The main contributions of our solution are threefold.
\begin{enumerate}
   \item
   We demonstrate how to increase the number of regions with memory efficiency by the novel inverted index structure. The efficient indexing contributes to high accuracy for approximate search.
   \item
   We describe how to encode data points  via a novel algorithm for high runtime/memory efficiency.
   \item
   Our evaluation shows that our system consistently and significantly outperforms state-of-the-art GPU- and CPU-based retrieval systems on both recall and efficiency on two public billion-scale benchmark datasets with single- and multi-GPU configurations.
\end{enumerate}

The rest of the paper is organized as follows. Section \ref{sec:s2} introduces related works on indexing with quantization methods. Section \ref{sec:s3} presents VLQ, our approach for approximate nearest neighbor (ANN)-based similarity search method. Section \ref{sec:s4} introduces the details of GPU implementation. Section \ref{sec:s5} provides a series of experiments, and compares the results to the state of the art.

\begin{figure}[htb]
    \begin{minipage}[t]{0.32\linewidth}
    \centering
    \includegraphics[width=.8\textwidth]{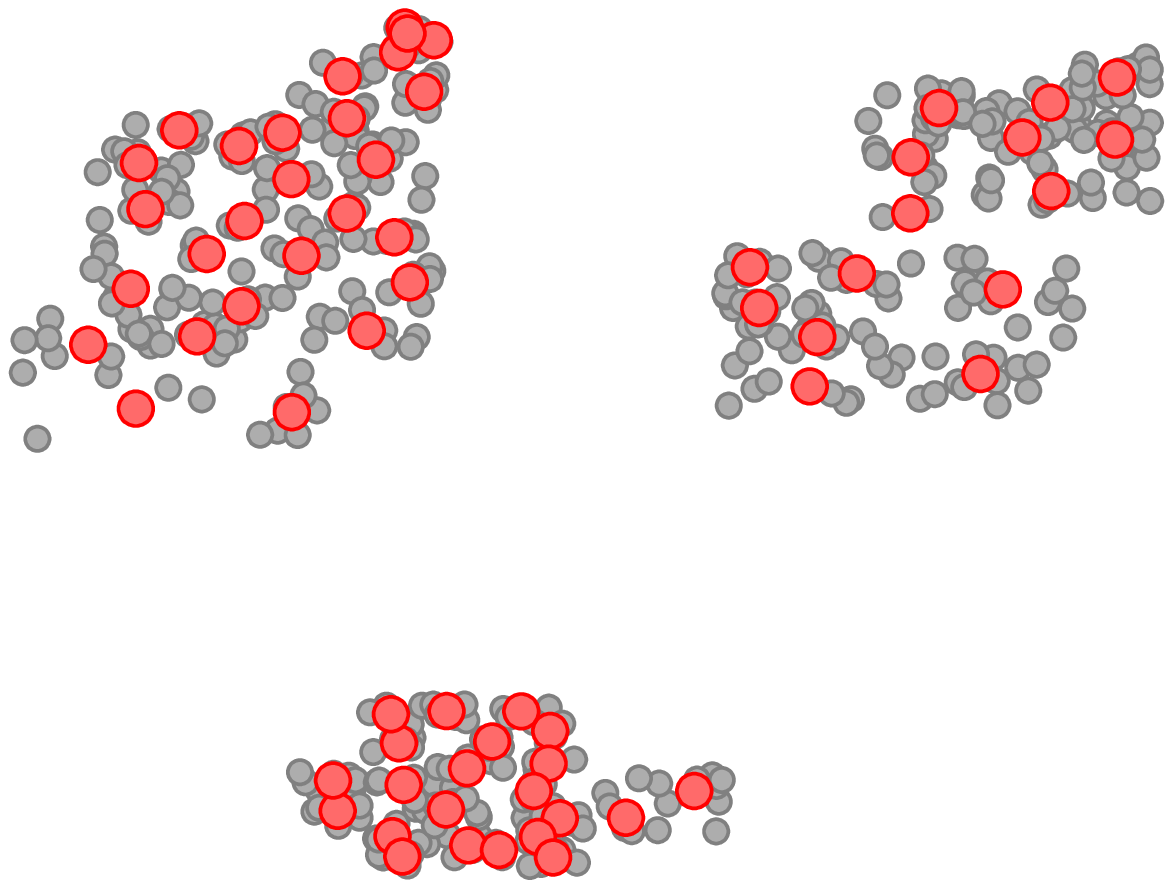}
    \caption*{\footnotesize{(a) Vector Quantization.}}
    \label{fig:side:a}
    \end{minipage}
    \begin{minipage}[t]{0.32\linewidth}
    \centering
    \includegraphics[width=.8\textwidth]{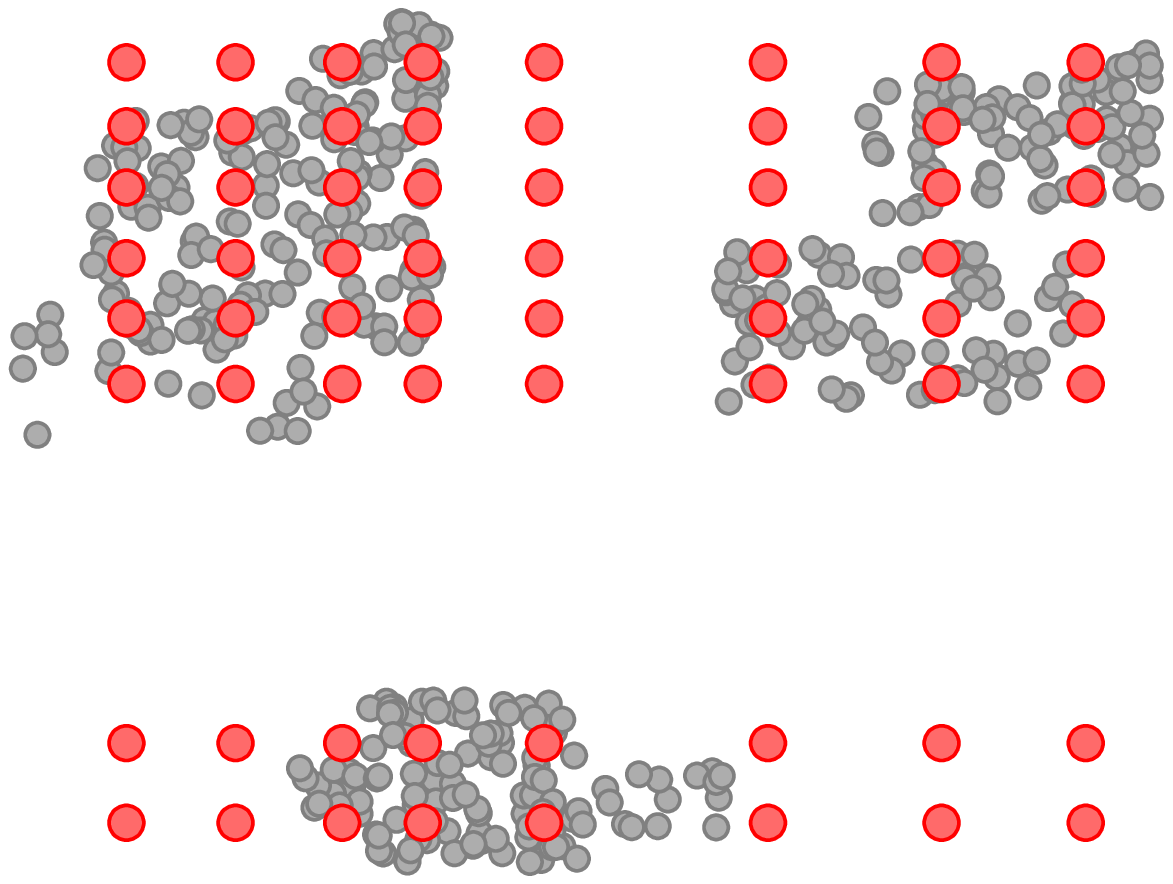}
    \caption*{\footnotesize{(b) Product Quantization.}}
    \label{fig:side:b}
    \end{minipage}
    \begin{minipage}[t]{0.32\linewidth}
    \centering
    \includegraphics[width=.68\textwidth]{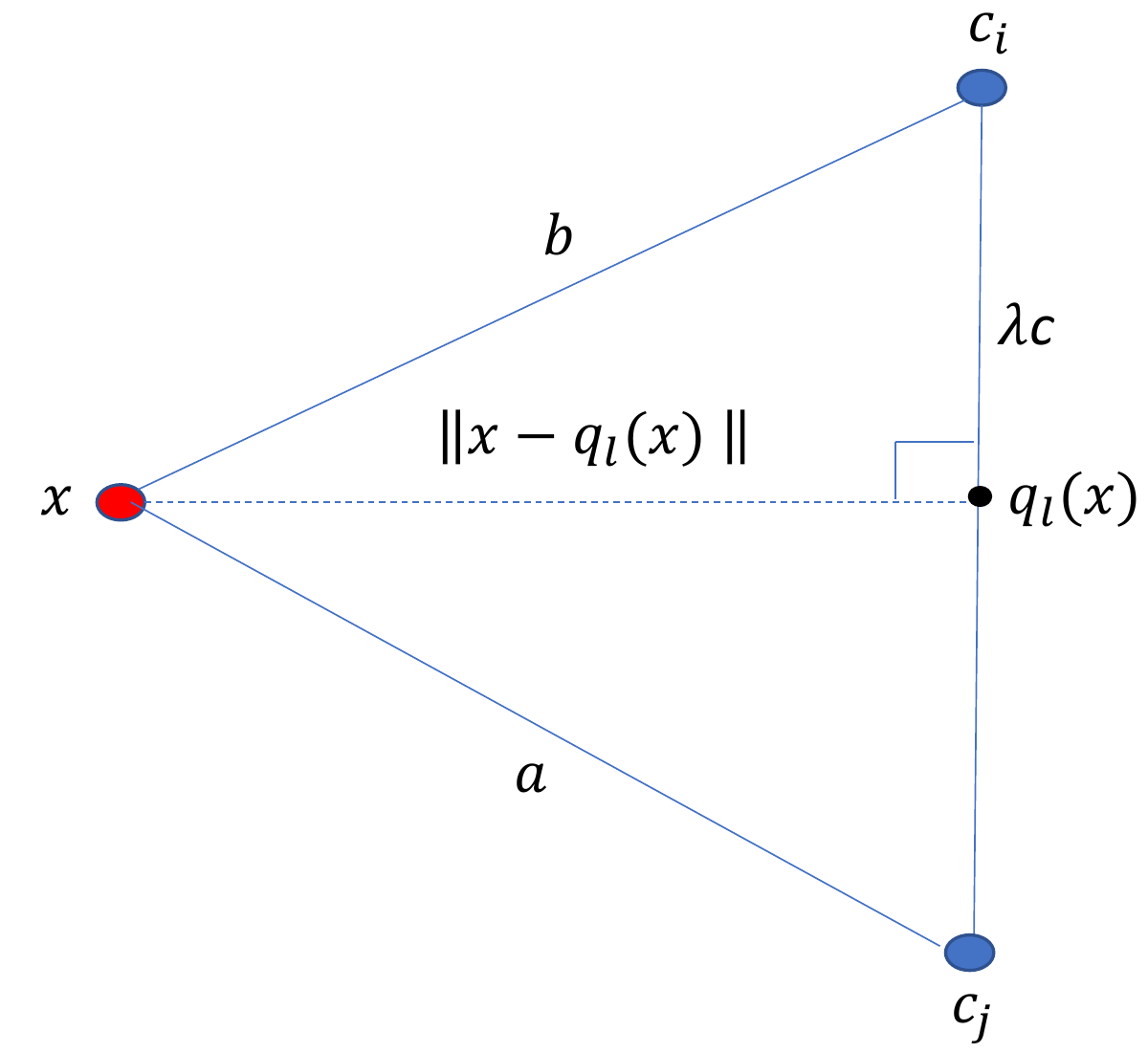}
    \caption*{\footnotesize{(c) Line Quantization.}}
    \label{fig:side:c}
    \end{minipage}
    \caption{Three different quantization methods. Vector and Product quantization methods are both with $k=64$ clusters. The red dots in plot (a) and (b) denote the centroids and the grey dots denote the dataset points in both plots. Vector quantization (a) maps the dataset points to the closest centroids. Product quantization (b) performs clustering in each subspace independently (here axes).
    In plot (c), a 2-dimensional point $x$ (red dot) is projected on line $l(c_i,c_j)$ with the anchor point $q_l(x)$ (black dot). The $a,b,c$ denote the values of $\parallel x- c_i\parallel^2$,$\parallel x- c_j\parallel^2$ and $\parallel c_i-c_j\parallel^2$ respectively. We use the parameter $\lambda$ to represent the value of $\parallel c_i- q_l(x)\parallel/c$ .  The anchor point $q_l(x)$ can be represented by $c_i,c_j$ and $\lambda$. The distance from $x$ to  $l(c_i,c_j)$ can be calculated by $a,b,c$ and $\lambda$.
    }
    \label{4p1}
\end{figure}

\begin{table}[htb]
\centering
\caption{Commonly used notations.}\label{tab:notations}
\begin{tabular}{lp{320pt}}
\toprule
Notation       & Description   \\
\midrule
 $x_i$, $D$    &  data points, their dimension and the number of data points     \\
 $\mathcal{X}$, $N$ & a set of data points and its size, $\mathcal{X} = \{x_1,\ldots,x_N\}\subset \mathbb{R}^D$\\
 $c, s, l(c,s)$  & centroids, nodes and edges      \\
 $m$           & encoding length \\
 $k$           & the number of first-level centroids      \\
 $n$           & the number of edges of each first-level centroid      \\
 $w_1$           & the number of first-layer nearest regions for a query\\
 $\alpha$      & the portion of the nearest of the $w\cdot n$  second-level regions \\
 $w_2$         & the number of second-level nearest regions for a query, $w_2 = \alpha\cdot w_1\cdot n$\\
 $\lambda$     & a scalar parameter for line quantization\\
 $r$           &displacement from data points to the approximate points\\
\bottomrule
\end{tabular}
\end{table}

\section{Related work}
\label{sec:s2}
In this section, we briefly introduce  some quantization methods and several retrieval systems related to our approach. Table~\ref{tab:notations} summarizes the common notations used throughout this paper. For example, we assume that $\mathcal{X}=\{x_1,\dots,x_N\}\subset\mathbb{R}^D$ is a finite set of $N$ data points of dimension $D$.

\subsection{Vector quantization (VQ)}
\label{s2.1}
In {vector quantization \cite{linde1980algorithm}} (Figure \ref{4p1} a), a quantizer is a function $q_v$ that maps a $D$-dimensional vector $x$ to a vector $q_v(x)\in C$, where $C$ is a finite subset of $\mathbb{R}^{D}$, of $k$ vectors. Each vector $c\in C$ is called a centroid, and $C$ is a codebook of size $k$.
We can use Lloyd iterations~\cite{lloyd1982least} to efficiently obtain a codebook $C$ on a subset of the dataset. For a finite dataset, $\mathcal{X}$, $q_v(x)$ induces quantization error $E$:

\begin{equation}
E= \sum_{x\in\mathcal{X}} \parallel x- q_v(x) \parallel^2. \label{e1}
\end{equation}

According to Lloyd's first condition, to minimize quantization error a quantizer should map vector $x$  to its nearest codebook centroid.

\begin{equation}
q_v(x)=\arg \min_{c\in C} \parallel x- c \parallel.
\end{equation}

Hence, the set of points $\mathcal{X}_{i}=\{x\in \mathbb{R}^{D} \mid q_v(x)=c_i\}$ is called a cluster or a region for centroid $c_i$.

The \textbf{inverted index structure} based on VQ~\cite{J2011Product} can split the dataset space into $k$ regions that correspond to the $k$ centroids of the codebook. Since the ratio of regions to centroids is $1\colon\!1$, it requires a large amount of space to store the $D$-dimensional centroids when $k$ is large. This would give a negative effect on the performance of the retrieval system. Our hierarchical index structure based on VLQ increase the ratio by $n$ times, i.e., $n$ times more regions can be generated by our indexing structure with the same number of centroids as the VQ based indexing structure.

\subsection{Product quantization (PQ)}
\label{s2.2}
Product quantization (Figure \ref{4p1} (b)) is an extension of vector quantization. Assuming that the dimension $D$ is a multiple of $m$, any vector $x\in\mathbb{R}^D$ can be regarded as a concatenation $(x^1,\cdots,x^m)$ of $m$ sub-vectors, each of dimension $D/m$. Suppose that $C^1,\cdots,C^m$ are $m$ codebooks of subspace $\mathbb{R}^{D/m}$, each owns $k$ $D/m$-dimensional sub-centroids. A codebook of a product quantizer $q_p$ is thus a Cartesian product of sub-codebooks.

 \begin{equation}
C=C^1\times\cdots\times C^m.
\end{equation}
Hence the codebook $C$ contains a total of $k^m$ centroids, each is a form of $c=(c^1,\cdots,c^m)$, where each sub-centroid $c^i\in C^i$ for $i\in\mathcal{ M}=\{1,\cdots,m\}$. A product quantizer $q_p$ should minimize the quantization error $E$ defined in Formula~\ref{e1}. Hence, for $x\in\mathbb{R}^{D} $, the nearest centroid in codebook $C$ is
 \begin{equation}
q_p(x)=(q_p^1(x^1),\cdots, q_p^m(x^m)),
\end{equation}
where $q^i$ is a sub-quantizer of $q$ and $q_p^i(x)$ is the nearest sub-centroid for sub-vector $x^i$, i.e.,  the nearest centroid $q_p(x)$ for $x$ is the concatenation of the nearest sub-centroids for sub-vector $x^i$.

The \textbf{inverted multi-index structure} (IMI) applies the idea of PQ for indexing and can generate $k^m$ regions with $m$ codebooks of $k$ sub-centroids each. The benefit of inverted multi-index is thus it can easily generate a much larger number of regions than that of VQ-based inverted index structure with moderate values of $m$ and $k$. The drawback of IMI is that it produces a lot of empty regions when the distributions of subspaces are not independent~\cite{Yandex2016Efficient}. This will affect the system's performance when handling datasets which have significant correlations between different subspaces, such as CNN-produced feature point dataset~\cite{Yandex2016Efficient}.

The PQ-based indexing structure later has been improved by OPQ~\cite{ge2013optimized} and LOPQ~\cite{Kalantidis2014Locally}. OPQ make a rotation on dataset points by a global $D\times D$ rotation matrix and LOPQ rotates the points which belong to the same cell by a same local $D\times D$ rotation matrix to minimize correlations between two subspaces~\cite{ge2013optimized}.  OPQ and LOPQ can both improve the indexing efficiency of PQ but slow down the query speed by a large margin as well.

Additionally, PQ can also be used to compress datasets. Typically each sub-codebook of PQ contains $256$ sub-centroids and each vector $x$ is mapped to a concatenation of $m$ sub-centroids  $(c^1_{j_1} ,\cdots,c^m_{j_m} )$, for $j_i$ is a value between $1$ and $256$. Hence the vector $x$ can be encoded into an $m$-byte code of sub-centroid index $(j_1 ,\cdots,j_m )$. With the approximate representation by PQ, the Euclidean distances between the query vector and the large number of compressed vectors can be computed efficiently. According to the ADC procedure~\cite{J2011Product}, the computation is performed based on lookup tables.
 \begin{equation}
\parallel y-x \parallel^2   \approx \parallel y-q_p(x) \parallel^2 = \sum_{i=1}^m \parallel y^i- c^i_{j_i} \parallel^2  \label{delta}
\end{equation}
where $y^i$ is the $i$th subvector of a query $y$. The Euclidean distances between the query sub-vector $y^i$ and each sub-centroids $c^i_{j_i}$ can be precomputed and stored in lookup tables that reduce the complexity of distance computation from $\mathcal{O}(\emph{D})$ to $\mathcal{O}(\emph{m})$. Due to the high compression quality and efficient distance computation approach, PQ is considered the top choice for compact representation of large-scale datasets\cite{ge2013optimized,Kalantidis2014Locally,Norouzi2013Cartesian,Baranchuk2018Revisiting,Babenko2014Improving}.

\subsection{Line quantization (LQ)}
Line quantization (LQ) \cite{wieschollek2016efficient} owns a codebook $C$ of $k$ centroids like VQ. As shown in Figure \ref{4p1} (c), with any two different centroids $c_{i},c_{j}\in C$, a line is formed and denoted by $l(c_{i},c_{j})$. A line quantizer $q_l$  quantizes a point $x$ to the nearest line as follows:

\begin{equation}
q_l(x)=\arg \min_{l(c_{i},c_{j})} d(x, l(c_{i},c_{j})),
\end{equation}
where $d(x, l(c_{i},c_{j}))$ is the Euclidean distance from $x$ to the line $l(c_{i},c_{j})$, and the set {$\mathcal{X}_{i,j}=\{x\in \mathbb{R}^{D} | q_l(x)=l(c_{i},c_{j})\}$} is called a cluster or a region for line $l(c_{i},c_{j})$. The squared distance $d(x, l(c_{i},c_{j}))$ can be calculated as following :
 \begin{equation}
  \begin{aligned}
d(x, l(c_{i},c_{j}))^2   &= (1-\lambda)\parallel x-c_{i} \parallel^2+(\lambda^2-\lambda) \parallel c_{j}-c_{i} \parallel^2\\&+ \lambda\parallel x-c_{j} \parallel^2  \label{delta1}
 \end{aligned}
\end{equation}
Because the values of $\parallel x-c_{j} \parallel^2 , \parallel x-c_{i} \parallel^2 , \parallel c_{j}-c_{i} \parallel^2$ can be pre-computed between $x$ and all centroids, Equation \ref{delta1} can be calculated efficiently. The anchor point of $x$ is represented by  $(1 - \lambda)\cdot c_{i} +\lambda\cdot c_{j}$, where $\lambda$ is a
scalar parameter that can be computed as following:
\begin{equation}
 \lambda =0.5 \cdot \frac{(\parallel x-c_{i} \parallel^2 + \parallel c_{j}-c_{i} \parallel^2 - \parallel x-c_{j} \parallel^2)} {\parallel c_{j}-c_{i} \parallel^2} \label{delta11}.
\end{equation}
 When $x$ is quantized to a region of $l(c_{i},c_{j})$, then the displacement of $x$ from $l(c_{i},c_{j})$ can be computed as following:
 \begin{equation}
r_{q_l}(x)=x-((1 - \lambda)\cdot c_{i} +\lambda\cdot c_{j}) \label{delta11a}.
\end{equation}

Here we regard  $l(c_{i},c_{j})$ and $l(c_{j},c_{i})$ as two different lines. So LQ-based indexing structrue can produce $k\cdot(k-1)$ regions with a codebook of $k$ centroids, The benefit of LQ-based indexing structure is that it can produce many more regions than that of VQ-based regions. However it is considerably more complicated to find the nearest line for a point $x$ when $k$ is large. So we use LQ as an indexing approach with a codebook of a few lines.

\begin{table}[htb]
\centering
\caption{A summary of current state-of-the-art retrieval systems based on quantization method. $N$ is the size of the dataset $\mathcal{X}$, $m$ is the number of sub-vectors in product quantizatino (PQ), $k$ is the size of the codebook, and $n$ is the number of second-level regions. In the last column of each row, the first term is the complexity for encoding, and the second term is the complexity for indexing.}\label{tab:sum}
 \resizebox{\textwidth}{!}{
\begin{tabular}{lllll}
\toprule
System       & Index structure  & Encoding & CPU/GPU & Space complexity \\
\midrule
Faiss~\cite{johnson2017billion}         & VQ         & PQ       & GPU   &  $\mathcal{O}(N\cdot m)+\mathcal{O}(k\cdot D)$ \\
Ivf-hnsw~\cite{Baranchuk2018Revisiting} & 2-level VQ & PQ       & CPU  &  $\mathcal{O}(N\cdot m)+\mathcal{O}(k\cdot (D+n))$   \\
Multi-D-ADC~\cite{babenko2012inverted}  & IMI (PQ)   & PQ       & CPU  &   $\mathcal{O}(N\cdot m)+\mathcal{O}(k\cdot (D+k))$ \\
\midrule
VLQ-ADC (our system)   & VLQ    & PQ     & GPU  &   $\mathcal{O}(N\cdot m)+\mathcal{O}(k\cdot (D+n))$ \\
\bottomrule
\end{tabular}
}
\end{table}

\subsection{The applications of VQ-based and PQ-based indexing structures for billion-scale dataset}
\label{s2.3}
In this subsection we introduce several billion-scale similarity retrieval systems that apply VQ- or PQ-based indexing structure and encoded by PQ, and discuss their strengths and weaknesses.

All the systems discussed below are best-performing, state-of-the-art systems for billion-scale high-dimensional ANN search. Their indexing structure and encoding method are summarized in Table~\ref{tab:sum}. Since all these systems employ the same encoding method based on PQ, we will mainly focus on their indexing structures in the discussions below.

\textbf{Faiss}~\cite{johnson2017billion} is a very efficient GPU-based retrieval approach, by realizing the idea of IVFADC \cite{J2011Product} on GPUs. Faiss uses the inverted index based on VQ \cite{sivic2003video} for non-exhaustive search and compresses the dataset by PQ. The inverted index of IVFADC owns a vector quantizer $q$ with a codebook of $k$ centroids. Thus there are $k$ regions for the data space. Each point $x\in\mathcal{X}$ is quantized to a region corresponding to a centroid by a VQ quantizer {$q_v$}. The displacement of each point from the centroid of a region it belongs to is defined as
 \begin{equation}
r_q(x)=x-q(x),
\end{equation}
where the displacement $r_q(x)$ is encoded by PQ with $m$ codebooks shared by all regions. For each region, an inverted list of data points is maintained, along with PQ-encoded displacements.

The search process of Faiss/IVFADC proceeds as follows:
\begin{enumerate}
  \item
  A query point $y$ is quantized to its $w$ nearest regions, extracting a list of candidates $\mathcal{L}_c\subset\mathcal{X}$ which
   have a high probability of containing the nearest neighbor.
  \item
  The displacement of the query point $y$ from the centroid of each sub-region is computed as $r_q(y)$.
   \item
   The distances between  $r_q(y)$ and  PQ-encoded displacements in $\mathcal{L}_c$ are then computed according to Formula~\ref{delta}.
    \item
   Sort the list $\mathcal{L}_c$ to be $\mathcal{L}_s$ based on the distances computed above. The first points in $\mathcal{L}_s$ are returned as the search result for query point $y$.
\end{enumerate}

\textbf{Ivf-hnsw}~\cite{Baranchuk2018Revisiting} is a retrieval system based on a two-level inverted index structure. Ivf-hnsw first splits the data space into $k$ regions like IVFADC. Then each region is further split into several sub-regions that correspond to $n$ sub-centroids. Each sub-centroid of a region can be represented by the centroid of the region and another centroid of a neighbor region. Assume that each region has $n$ neighbor regions, thus each region can be split into $n$ regions. Each data point is first quantized to a region and then further quantized to a sub-region of the region. The displacement of each point from the sub-centroid of a sub-region it belongs to is encoded by PQ. An inverted list of data point is maintained for each sub-regions similar to IVFADC.

The search process of Ivf-hnsw proceeds as follows:
\begin{enumerate}
  \item
  A query point $y$ is quantized to its $w$ first-level nearest regions, giving $w\cdot n$ sub-regions.
  \item
   Among the $w\cdot n$ sub-regions, $y$ is secondly quantized to $0.5\cdot w\cdot n$ nearest sub-regions, generating a list of candidates $\mathcal{L}_c\subset\mathcal{X}$.
  \item
  The displacement of the query point $y$ from the sub-centroid of each sub-region is computed as $r_q(y)$.
   \item
   The distances between  $r_q(y)$ and  PQ-encoded displacements in $\mathcal{L}_c$ are then computed according to Formula \ref{delta}.
    \item
   The re-ordering process of Ivf-hnsw is similar to IVFADC/Faiss.
\end{enumerate}

\textbf{Multi-D-ADC}~\cite{babenko2012inverted} is based on the inverted multi-index which is currently the state-of-the-art indexing method for high-dimensional large-scale datasets.  An inverted multi-index of Multi-D-ADC usually owns a product quantizer with two sub-quantizers $q^1, q^2$ for subspace $\mathbb{R}^{D/2}$, each of $k$ sub-centroids. A region in the D-dimensional space is now  a Cartesian product of two corresponding subspace regions. So the IMI can produce $k^2$ regions. For each point $x=(x^1,x^2)\in\mathcal{X}$, sub-vectors $x^1,x^2 \in \mathbb{R}^{D/2}$  are separately quantized to subspace regions of $q^1(x^1),q^2(x^2)$ respectively, and $x$ is then quantized to the region of $(q^1(x^1),q^2(x^2))$ . The displacement of each point $x$ from the centroid $(q^1(x^1),q^2(x^2))$  is also encoded by PQ,
and an inverted list of points is again maintained for each region.

The search process of Multi-D-ADC proceeds as follows:
\begin{enumerate}
  \item
  For a query point $y=(y^1,y^2)$, The Euclidean distances of each of sub-vectors $y^1,y^2$ to all sub-centroids of $q^1,q^2$ are computed respectively. The distance of $y$ to a region can be computed according to Formula \ref{delta} for $m=2$.
  \item
 Regions are traversed in ascending order of distance to $y$ by the multi-sequence algorithm~\cite{babenko2012inverted} to generate a list of candidates $\mathcal{L}_c\subset\mathcal{X}$.
  \item
  The displacement of the query point $y$ from the centroid $(c^1,c^2)$ of each region is computed as $r_q(y)$ as well.
 \item
 The re-ordering process of Multi-D-ADC is similar to IVFADC/Faiss.
\end{enumerate}

The VQ-based indexing structure requires a large full-dimensional codebook to produce regions when $k$ is large. The PQ-based indexing structure are not suitable for all datasets, especially for those produced by convolutional neural networks (CNN) \cite{Baranchuk2018Revisiting}. The novel VQ-based indexing structure proposed by Ivf-hnsw can produce more regions than the prior VQ-based indexing structure. However its performance on the codebook of small size is not good enough. We will discuss that in Sec.\ref{sec:s5}. In comparison, our indexing structure is efficient with a small size of codebook which can accelerate query speed and at the same time is suitable for any dataset irrespective of the presence/absence of correlations between subspaces.

\section{The VLQ-ADC System}
\label{sec:s3}
In this section we introduce our GPU-based similarity retrieval system, VLQ-ADC, that contains a two-layer hierarchical indexing structure based on vector and line quantization and an asymmetric distance computation method. VLQ-ADC incorporates a novel index structure that can index the dataset points efficiently (Sec.~\ref{sec:vlq}). The indexing and encoding process will be presented in Sec.~\ref{sec:algo}, and the querying process is discussed in Sec.~\ref{sec:query}.

\begin{figure*}[!htb]
    \begin{minipage}[t]{0.48\linewidth}
    \centering
    \includegraphics[width=\textwidth]{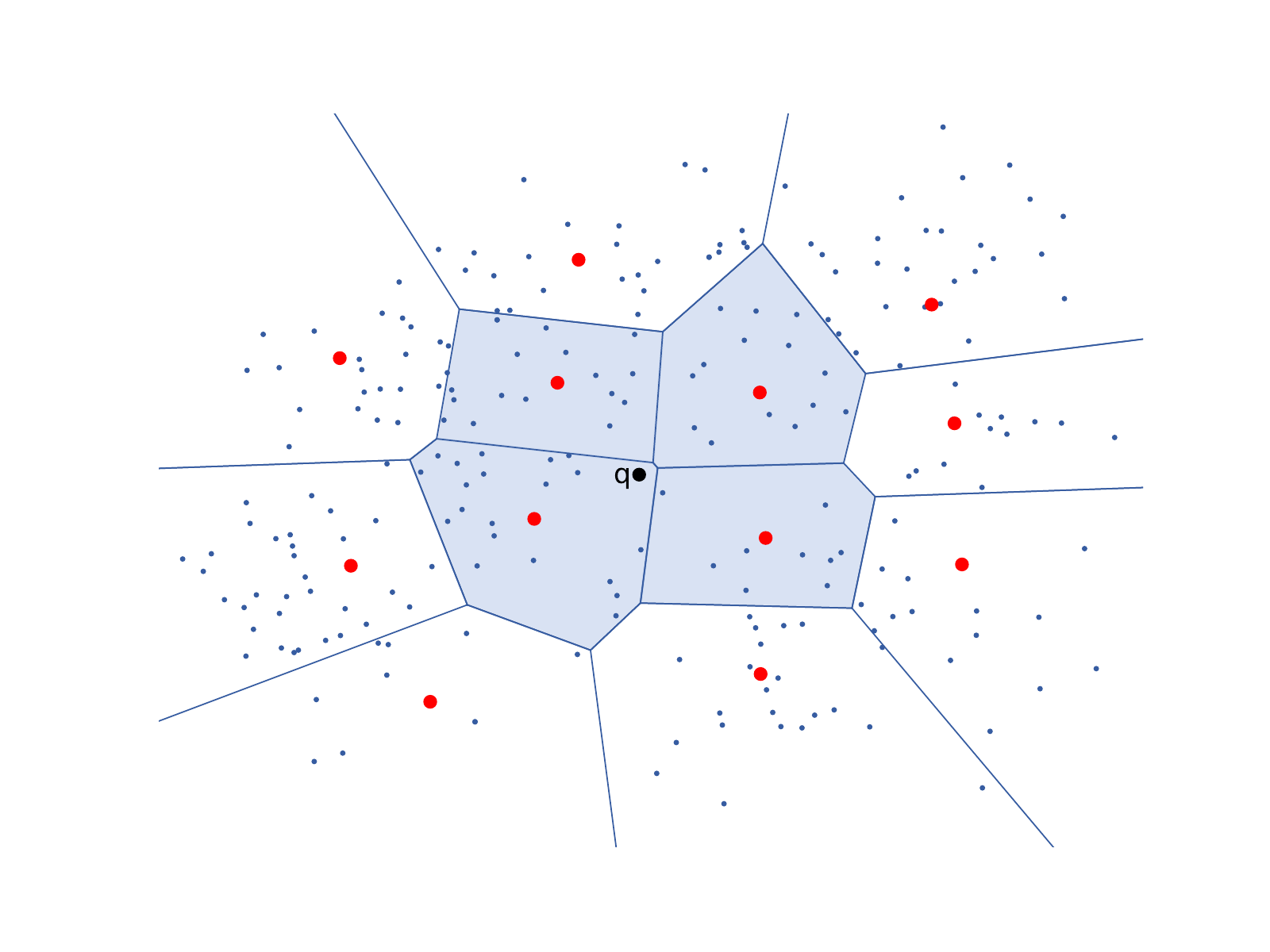}
    \caption*{\footnotesize{VQ-based indexing structure}}
    \label{fig:side:a}
    \end{minipage}
    \begin{minipage}[t]{0.48\linewidth}
    \centering
    \includegraphics[width=\textwidth]{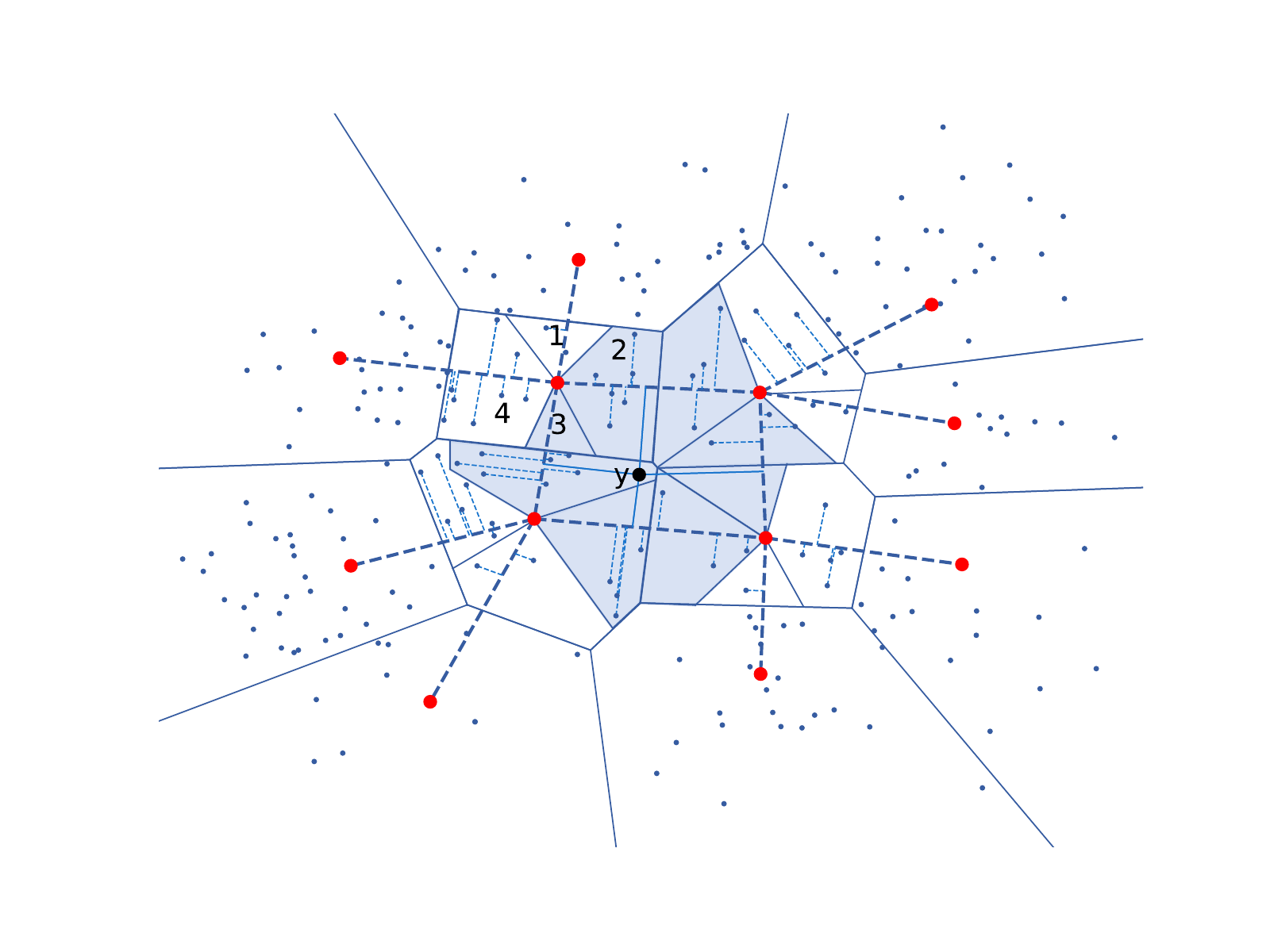}
    \caption*{\footnotesize{VLQ-based indexing structure}}
    \label{fig:side:b}
    \end{minipage}
     \caption{A comparison of the indexing structure and search process of the VQ-based indexing structure (\textbf{left}) and our VLQ-based indexing structure (\textbf{right}) on data points (small blue dots) of dimension 2 ($D=2$). The large red dots denote the (first-level) same cell centroids in both figures. \textbf{Left:} The 4 shaded areas in the left figure represent the first-level regions, one for each centroid, and they make up the areas that need to be traversed for the query point $q$. \textbf{Right:} For each centroid in the right figure, $n=4$ nearest neighboring centroids are found. Thus the $n$-NN graph consists of all the centroids and the edges (thick dashed lines) between them. Each first-level region in the right figure consists of 4 second-level regions, each of which represent the data points closet to the corresponding edge in the $n$-NN graph as denoted by the line quantizer $q_l$. Given the query point $q$ and parameter $\alpha=0.5$, only half of the second-level subregions (shaded in blue) need to be traversed. As can be seen, VLQ allows search to process substantially smaller regions in the dataset than a VQ-based approach.}
    \label{fig:vlq}
\end{figure*}

Comparing with the existing systems above, One major advantage of our system is that our indexing structure can generate shorter and more accurate candidate list for the query point, which will accelerate query speed by a large margin. Another advantage of our system is that the improved asymmetric distance computation method base on PQ encoding method provide a higher search accuracy. In the remainder of this section we will use Figure~\ref{fig:vlq} to illustrate our framework. We recall that commonly used notations are summarized in Table~\ref{tab:notations}.

\subsection{The VLQ-based index structure}\label{sec:vlq}
For billion-scale datasets with a moderate number of regions (e.g., $2^{16}$) produced by vector quantization (VQ), the number of data points in most regions is too large, which negatively affects search accuracy. To alleviate this problem, we propose a hierarchical indexing structure. In our structure, each list is split into several shorter lists, i.e., each region is divided into several subregions, using line quantization (LQ).

Our indexing structure is a two-layer hierarchical structure which consists of two levels of quantizers. The first level contains a vector quantizer $q_v$ with a codebook of $k$ centroids. The vector quantizer $q_v$ partitions the data point space $\mathcal{X}$ into $k$ regions. The second level contains a line quantizer  $q_l$ with an $n$-nearest neighbor ($n$-NN) graph. The $n$-NN graph is a directed graph in which nodes are first-level centroids and edges connect a centroid to its $n$ nearest neighbors. In each first-level region, the line quantizer $q_l$ then quantizes each data point to the closest edge in the $n$-NN graph, thus splitting the region into $n$ second-level regions.

As an example, in the right side of Figure~\ref{fig:vlq}, given $n=4$, the top left first-level region is further divided into 4 subregions by $q_l$, enclosed by solid lines and denoted 1, 2, 3, and 4. Each subregion contains all the data points that are closest to a given edge of the $n$-NN graph, as calculated by the line quantization $q_l$.

\textbf{Training the codebook}. We use Lloyd iteration in the fashion of the Linde-Buzo-Gray algorithm \cite{linde1980algorithm} to obtain the codebook of the VQ quantizer $q_v$. The $n$-NN graph is then built on the centroids of the codebook.

\textbf{Memory overhead of indexing structure}. One advantage of our indexing structure is its ability to produce substantially more subregions with little additional memory consumption. Same as VQ, our first layer codebook needs $k\cdot D \cdot sizeof(\texttt{float})$ bytes. In addition, for second-level indexing, for each of the $k$ first-layer centroids, the $n$-NN graph only needs to store (1) the indices of its $n$ nearest neighbors and (2) the distances to its $n$ nearest neighbors, which amounts to $k\cdot n \cdot (sizeof(\texttt{int})+sizeof(\texttt{float}))$ bytes. Note we do not need to store the full-dimensional points. For a typical values of $k=2^{16}$ centroids and $n=32$ subcentroids, the additional memory overhead for storing the graph is $2^{16}\cdot 32\cdot (32+32)$ bits (16 MB), which is acceptable for billion-scale datasets.

One way to produce the subregions is by utilizing vector quantization (VQ) again in each region. However, that would require storing full-dimensional subcentroids and thus consume too much additional memory. For the same configuration ($k=2^{16}$ centroids and $n=32$ subcentroids) and a dimension of $D=128$, the additional memory overhead for a VQ-based hierarchical indexing structure would be $2^{16}\cdot 32\cdot128 \cdot sizeof(\texttt{float}) $ additional bits (1,024 MB). As can be seen, our VLQ-based hierarchical indexing structure is substantially more compact, only consuming $1/64$ of the memory required by a VQ-based approach for the second-level codebook. 

We note that the PQ-based indexing structure requires $\mathcal{O}(k\cdot (D + k))$ memory to maintain
the indexing structure (Table~\ref{tab:sum}), which is memory inefficient as it is quadratic in $k$. This is a limitation of PQ-based indexing structure. In contrast, the space complexity of our hierarchical indexing structure is $\mathcal{O}(k\cdot(D+n))$, where typically $n \ll k$ ($n$ is much smaller than $k$), hence making our index much more memory efficient.

\begin{algorithm}[htb]
\caption{VLQ-ADC batch indexing process}
\label{alg:A}
\begin{algorithmic}[1] 
    \Function {Index}{$[x_1 ,\ldots,x_{N} ]$}
       \For{$t \gets 1 : N $}
            \State  $x_t \mapsto q_v(x)=\mathop{\arg\min}_{c\in C}\parallel x_{t}-c\parallel^2$ \hfill \texttt{// VQ}
             \State $S_i = n$-$\mathop{\arg\min}_{c\in C}\parallel c-c_i\parallel^2$ \hfill \texttt{// Construct the $n$-NN graph}
            \State $x_t \mapsto q_l(x)=\arg \min_{l(c_{i}, s_{ij}),s_{ij}\in S_i} d(x, l(c_{i},s_{ij}))$ \hfill \texttt{// LQ}

        \EndFor
    \EndFunction
\end{algorithmic}
\end{algorithm}

\begin{algorithm}[htb]
\caption{VLQ-ADC batch encoding process}
\label{alg:B}
\begin{algorithmic}[1] 
     \Function {Encode}{$[x_1 ,\ldots,x_{N} ]$}
       \For{$t \gets 1 : N$}
            \State $r_{q_l}(x_t)= x_t-((1 - \lambda_{ij})\cdot c_i +\lambda_{ij}\cdot {s_{ij}})$ \hfill\texttt{// Equation~\ref{delta131}}
            \State let $r_t = r_{q_l}(x_t)$
            \hfill \texttt{// displacement}
            \State $r_t=[r_ t^1,\ldots,r_t^m]$ \hfill\texttt{// divide $r_t$ into $m$ subvectors}
            \For{$p \gets 1 : m $}
            \State $r_t^p \mapsto c_{j_p} =  \arg\min_{c_{j_p}\in C^p}\parallel r_{t}^p-c_p\parallel^2 $
            \EndFor
             \State $Code_t = (j_1,\ldots,j_m)$
        \EndFor
    \EndFunction
\end{algorithmic}
\end{algorithm}

\subsection{Indexing and encoding}
\label{sec:algo}

In this subsection, we will describe the indexing and encoding process and summarize both processes in Algorithm \ref{alg:A} and \ref{alg:B} respectively. 

For our two-level index structure, the indexing process comprises two different quantization procedures, one for each layer. Similar to the IVFADC scheme, each dataset point is quantized by the vector quantizer $q_v$ to the first-level regions surrounded by the dotted lines in Figure~\ref{fig:vlq}. These regions form a set of inverted lists as search candidates.

We describe the second-level indexing process as follows. Let {$\mathcal{X}^{i}$ be a region of $\{x_1,\ldots,x_l\}$} that corresponds to a centroid $c_i$, for $i\in\{1,\ldots,k\}$. In constructing the $n$-NN graph, let $S_i = \{s_{i1},\ldots,s_{in}\}$ denote the set of the $n$ centroids closest to $c_i$ and $l(c_i,s_{ij})$ denote an edge between $c_i$ and $s_{ij}$, for $j\in\{1,\ldots,n\}$. The points in $\mathcal{X}^{i}$ are quantized to the subregions by a line quantizer $q_l$ with a codebook $\mathcal{E}_{i}$ of $n$ edges $\{l(c_i,s_{i1}),\ldots,l(c_i,s_{in})\}$. Thus the region $\mathcal{X}^{i}$ is split into $n$  subregions  $\{\mathcal{X}^i_{1},\ldots,\mathcal{X}^i_{n}\}$ and each point $x \in \mathcal{X}^{i}$ is quantized to a second-level subregion $\mathcal{X}^i_{j}$. So the entire space $\mathcal{X}$ are divided into $k\times n$ second-level subregions.
\begin{equation}
 \mathcal{X}^i_j=\{x\in \mathcal{X}^{i} \mid q_l(x)=l(c_i,s_{ij})\}\text{, for all $i\in \{1\ldots k\}$}
\label{delta12}
\end{equation}

Each data point in the dataset $\mathcal{X}$ is assigned to one of the $k\cdot n$ cells. When the data point $x$ is quantized to the sub-region of edge $l(c_{i},s_{ij})$,  according to the Equation \ref{delta11a} and \ref{delta11} the displacement of $x$ from the corresponding anchor point can be computed as following:
 \begin{align}
r_{q_l}(x)&=x-((1 - \lambda_{ij})\cdot c_{i} +\lambda_{ij}\cdot s_{ij}) \label{delta131}, \text{ where}\\
 \lambda_{ij} &=-0.5 \cdot \frac{(\parallel x-s_{ij} \parallel^2 - \parallel x-c_{i} \parallel^2 - \parallel s_{ij}-c_{i} \parallel^2)} {\parallel s_{ij}-c_{i} \parallel^2} \label{delta13}.
\end{align}.

As shown in Algorithm \ref{alg:B}, the value of $r_{q_l}(x)$ is first computed by Equation \ref{delta131} and encoded into $m$ bytes using PQ \cite{J2011Product}. The PQ codebooks are denoted by $C^1,\ldots,C^m$, each containing 256 sub-centroids. The vector $r_{q_l}(x)$ is mapped to a concatenation of $m$ sub-centroids  $(c^1_{j_1} ,\cdots,c^m_{j_m} )$, for $j_i$ is a value between $1$ and $256$. Hence the vector $r_{q_l}(x)$ is encoded into an $m$-byte code of sub-centroid index $(j_1 ,\cdots,j_m )$. In Figure \ref{4p1}(c), we assume that $c_i$ is the closest centroid to $x$ and can observe that the anchor point of each point $x$ is closer to $x$ than $c_i$. So the dataset points can be encoded more accurately with the same code length. This will improve the recall rate of search, as can be seen in our evaluation in Section~\ref{sec:s5}.

From {Equation \ref{delta13}, the value of $\lambda_{ij}$ for each point can be computed. it is a \texttt{float} type value and requires 4 bytes for each data point. To further improve memory efficiency, we quantize it into 256 values and encode it by a byte. Empirically we find that the encoded $\lambda_{ij}$ still exhibits high recall rates.

\subsection{Query}
\label{sec:query}
One important advantage of our indexing structure is that at query time, a specific query point only needs to traverse a small number of cells whose edges are closest to the query point, as shown in the right side of Figure~\ref{fig:vlq}. There are three steps for query processing: (1) region traversal, (2) distance computation and (3) re-ranking.

\subsubsection{Region traversal}
\label{s3.3.1}
The region traversal process consists of two steps: first-level regions traversal and second-level regions traversal. During first-level regions traversal, a query point $y$ is quantized to its $w_1$ nearest first-level regions, which correspond to $w_1\cdot n$ second-level regions produced by quantizer $q_v$. The subregions traversal is performed within only the $w_1\cdot n$ second-level regions. Moreover, $y$ is quantized again to $w_2$ nearest second-level regions by quantizer $q_l$. Then the candidate list of $y$ is formed by the data points only within the  $w_2$ nearest second-level regions. Because  the  $w_2$ second-level regions is obviously smaller than the $w_1$ first-level regions, the candidate list produced by our VLQ-based indexing structure is shorter than that produced by the VQ-based indexing structure. This will result in a faster query speed.

We use parameter $\alpha$ to determine the percentage of $w_1\cdot n$ second-level regions to be traversed give a query, such that $w_2=\alpha\cdot w_1 \cdot n$. We conduct a series of experiments in Section \ref{sec:s5} to discuss the performance of our system with different values of $\alpha$.

\subsubsection{Distance computation}
\label{s3.3.2}
 Distance computation is a prerequisite condition for re-ranking. In this section, we describe how to compute the approximate distance between a query point $y$ to a candidate point $x$. According to \cite{J2011Product}, the distance from $y$ to $x$ can be evaluated by asymmetric distance computation (ADC) as follows:
 \begin{equation}
 \parallel y-q_1(x)-q_2(x-q_1(x)) \parallel^2 \label{delta2}
\end{equation}
where $q_1(x) =(1 - \lambda_{ij})\cdot c_{i} +\lambda_{ij}\cdot{s_{ij}}$  and $q_2(\cdots)$ is the PQ approximation of the $x_i$ displacement.

Expression \ref{delta2} can be further decomposed as follows~\cite{Babenko2014Improving}:
 \begin{equation}
  \begin{aligned}
 \parallel y- q_1(x)& \parallel^2+\parallel q_2(\cdots) \parallel^2 + 2\langle q_1(x), q_2(\cdots)\rangle -\\&2\langle y, q_2(\cdots)\rangle \label{delta3}.
 \end{aligned}
\end{equation}
where  $\langle \cdot,\cdot \rangle$ denotes the inner product between two points.


If $l(c_i,s_{ij})$ is the closest edge to $x$, i.e., $q_1(x)=(1 - \lambda_{ij})c_i +\lambda_{ij}s_{ij}$, Expression \ref{delta3} can be transformed in the following way:
 \begin{equation}
  \begin{aligned}
 &\underbrace {\parallel y- ((1 - \lambda_{ij})c_i +\lambda_{ij}s_{ij}) \parallel^2}_{\text{term1}}+\underbrace {\parallel q_2(\cdots) \parallel^2}_{\text{term2}} +\\& 2(1 - \lambda_{ij})\underbrace {\langle c_i , q_2(\cdots)\rangle}_{\text{term3}}+ 2\lambda_{ij}\underbrace {\langle s_{ij}, q_2(\cdots)\rangle}_{\text{term4}} -\underbrace {2\langle y, q_2(\cdots)\rangle}_{\text{term5}} \label{delta4}.\\
 \end{aligned}
\end{equation}

According to Equation \ref{delta1}, term1 in Expression \ref{delta4} can be computed in the following way:
\begin{equation}
 \begin{aligned}
\parallel y- &((1 - \lambda_{ij})c_i +\lambda_{ij}s_{ij}) \parallel^2=(1-\lambda_{ij}) \underbrace {\parallel y- c_i  \parallel^2}_{\text{term6}}+\\&(\lambda_{ij}^2 -\lambda_{ij})\underbrace {\parallel c_i - s_{ij} \parallel^2}_{\text{term7}}+ \lambda_{ij} \underbrace {\parallel y-s_{ij} \parallel^2 }_{\text{term8}}. \label{delta5}
\end{aligned}
\end{equation}

In Expression \ref{delta4} and Equation \ref{delta5}, some computations can be done in advance and stored in lookup table as follows:
\begin{itemize}
  \item All of term2, term3, term4 and term7 are independent of the query. They can be precomputed from the codebooks. Term2 is the squared norm of the displacement approximation and can be stored in a table of size $256\times m$. Term7 is the square of the length of the edge that the point $x$ belongs to and is already computed in the codebook learning process. Term3 and term4 are scalar products of the PQ sub-centroids and the corresponding first-level centroid subvectors and can be stored in a table of size $k\times256\times m$.
  \item Term6 and term8 are the distances from the query point to the first-layer centroids. They are the by-product of first-layer traversal.
  \item Term5 is the scalar product of the PQ sub-centroids and the corresponding query subvectors and can be computed independently before the search. Its computation costs $256\times D$ multiply-adds \cite{johnson2017billion}.
\end{itemize}

The proposed decomposition is used to simplify the distance computation. With the lookup tables, the distance computation only requires $256\times D$ multiply-adds and $2\times m$ lookup-adds. In comparison, the classic IVFADC distance computation requires $256\times D$ multiply-adds and $m$ lookup-adds \cite{johnson2017billion}. The additional $m$ lookup-adds in our framework improves the distance computation accuracy with a moderate increase of time overhead. We will discuss this trade-off in detail in Section \ref{sec:s5}.

\subsubsection{Re-ranking}
\label{s3.3.3}

Re-ranking is a step of re-sorting the candidate list of data points according to the distances from candidate points to the query point. It is the last step of the query process. The purpose of re-ranking is to find out the nearest neighbours to the query point among the candidate points by distance comparing.We apply the fast sorting algorithm of \cite{johnson2017billion} to our re-ranking step. Due to the shorter candidate list and more accurate approximate distances, the re-ranking step of our system is both faster and more accurate than that of Faiss. 
\begin{algorithm}[!htb]
        \caption{VLQ-ADC batch search process}
        \label{alg:C}
        \begin{algorithmic}[1] 
            \Function {Search}{$[y_1 ,\cdots,y_{n_q} ], \mathcal{L}_1,\cdots, \mathcal{L}_{k\times n}$}
               \For{$t\gets 1 : n_q $}
                    \State $C_t\gets w_1$-$\mathop{\arg\min}_{c\in C}\parallel  y_{t}-c\parallel^2$\label{ln:wNN}
                    \State $\textit{L}^t_{LQ}\gets w_2$-$\arg\min_{c_i\in C_t, s_{ij}\in S_{i}}\parallel y-(1 - \lambda_{ij})\cdot c_i -\lambda_{ij}\cdot s_{ij}\parallel^2$ \hfill\texttt{// described in Sec.~\ref{s3.3.1}}\label{ln:w1NN}
                    \State Store values of $\parallel  y_{t}- c \parallel^2$
                \EndFor
                \For{$t\gets 1 : n_q $}
                   \State $\textit{L}_t \gets []$
                    \State Compute $\langle y_{t}, q_2(\cdots)\rangle$ \hfill\texttt{// See term5 in Equation \ref{delta4}}
                     \For{$\textit{L}$ in $\textit{L}^t_{LQ} $}
                      \For{$t'$ in $\mathcal{L}_\textit{L} $}
                      \State\texttt{// distance evaluation described in Sec.~\ref{s3.3.2}}
                      \State $d \gets \parallel y_{t}-q_1(x_{t'})-q_2(x_{t'}-q_1(x_{t'}))\parallel^2$
                      \State Append ($d$; $\textit{L}$; $j$) to $\textit{L}_t$
                      \EndFor
                      \EndFor
                \EndFor
                \State $R_t\gets $ K-smallest distance-index pairs ($d$, $t'$) from $\textit{L}_t$ \hfill\texttt{// Re-ranking}
                \State \Return{$R_t$}
            \EndFunction

        \end{algorithmic}
    \end{algorithm}

\section{GPU Implementation}
\label{sec:s4}
One advantage of our VLQ-ADC framework is that it is amenable to implementations on GPUs. It is mainly because our searching and distance computing algorithm that applied during query can be efficiently parallelized on GPUs. In this work we have implemented our framework in CUDA.

There are three different levels of granularity of parallelism on GPU: threads, blocks and grids. A block is composed of multiple threads, and a grid is composed of multiple blocks. Furthermore, there are three memory types on GPU. Global memory is typically $4$--$32$ GB in size with $5$--$10\times$ higher bandwidth than CPU main memory \cite{johnson2017billion}, and can be shared by different blocks. Shared memory is similar to CPU L1 cache in terms of speed and is only shared by threads within the same block. GPU
register file memory has the highest bandwidth and the size of register file memory on GPU is much larger than that on CPU~\cite{johnson2017billion}.

VLQ-ADC is able to utilize GPU efficiently for indexing and search. For example, we use blocks to process $D$-dimensional query points and the threads of a block to traverse the inverted lists. We use global memory to store the indexing structures and compressed dataset that is shared by all blocks and grids, and load part of lookup tables in the shared memory to accelerate distance computation. As the GPU register file memory is very large, we store structured data in the register file memory to increase the performance of  the sorting algorithm.

Algorithm \ref{alg:C} summarizes our search process that is implemented on GPU. We use four arrays to store the information of the inverted index lists. The first array stores the length of
each index list, the second one stores the sorted vector IDs of each list, and the third the fourth store the corresponding codes and $\lambda$ values of each list respectively. For an NVIDIA GTX Titan X GPU with a 12GB of RAM, we load part of the dataset indexing structure in the global memory for different kernels, i.e., region size, data points compressed codes and $\lambda$ values of each list. A kernel is the unit of work (instruction stream with arguments) scheduled by the host CPU and executed by GPUs~\cite{johnson2017billion}. We load the vector IDs on the CPU side, because vector IDs are resolved only if re-ranking step determines K-nearest membership. This lookup produces a few sparse memory reads in a large array, thus the IDs stored on CPU can only cause a tiny performance cost.

Our implementation makes use of some basic functions from the Faiss library, including matrix multiplication and the K-selection algorithm to improve the performance of our approach\footnote{The source code will be released upon publication.}.

\paragraph{K-selection} The K-selection algorithm is a high-performance GPU-based sorting method proposed by Faiss \cite{johnson2017billion} and GSKNN \cite{yu2015performance}. The K-selection keep intermediate data in the register file memory. It exchanges register data using the warp shuffle instruction, enabling warp-wide parallelism and storage. The warp is a 32-wide vector of GPU threads, each thread in the warp has up to 255 32-bit registers in a shared register file. All the threads in the same warp can exchange register data using the warp shuffle instruction.

\paragraph{List search} We use two kernels for inverted list search. The first kernel is responsible for quantizing each query point to $w_1$ nearest first-level regions (line \ref{ln:wNN} in Algorithm~\ref{alg:C}). The second kernel is responsible for finding out the $w_2$ nearest second-level regions for the query point (line \ref{ln:w1NN} in Algorithm~\ref{alg:C}). The distances between each query point and its $w$ nearest centroids are stored for further calculation. In the two kernels, we use a block of threads to process one query point, thus a batch of  $n_q$  query points can be processed concurrently.

\paragraph{Distance computation and re-ranking} After the inverted lists $\mathcal{L}_\textit{i}$ of each query point are collected, there are up to $n_q\times w_2 \times \max|\mathcal{L}_\textit{i}|$ candidate points to process. During the distance computation and re-ranking process, processing all the query points in a batch yields high parallelism, but can exceed available
GPU global memory. Hence, we choose a tile size $t_q<n_q$ based on amount of available memory to reduce memory overhead, bounding its complexity by $\mathcal{O}(t_q \times w_2\times max|\mathcal{L}_i|)$.

We use one kernel to compute the distances from each query point to the candidate points according to Expression \ref{delta4}, and sort the distances via the K-selection algorithm in a separate kernel. The lookup tables are stored in the global memory. In the distance computation kernel, we use a block to scan all $w_1$ inverted lists for a single query point, and the significant portion of the
runtime is the $2\times w_2\times m$  lookups in the lookup tables and the linear scanning of the $\mathcal{L}_i$ from global memory.

In the re-ranking kernel, we refer to Faiss by using a two-pass K-selection. First reduce $t_q \times w_2\times max|\mathcal{L}_\textit{i}|)$ to $t_q \times \tau \times K$ partial results, where $\tau $ is some subdivision factor, then the partial results are reduced again via k-selection to the final $t_q \times K$ results.

Due to the limited amount of GPU's memory, if an index instance with long encoding length cannot fit in the memory of a single GPU, it cannot be processed one the GPU efficiently. Our framework supports multi-GPU parallelism to process indexing instance of a long encoding length.
For $b$ GPUs, we split the index instance into $b$ parts, each of which can fit in the memory of a single GPU. We then process the local search of $n_q$ queries on each GPU, and finally join the partial results on one GPU. Our multi-GPU system is based on MPI, which can be easily extended to multiple GPUs on multiple servers.

\section{Experiments and Evaluation}
\label{sec:s5}

In this section, we evaluate the performance of our system VLQ-ADC and compare it to three state-of-the-art billion-scale retrieval systems that are based on different indexing structures and implemented on CPUs or GPUs: Faiss~\cite{johnson2017billion}, Ivf-hnsw~\cite{Baranchuk2018Revisiting} and Multi-D-ADC~\cite{babenko2012inverted}. All the systems are evaluated on the standard metrics: accuracy and query time, with different code lengths. All the experiments are conducted on a machine with two 2.1GHz Intel Xeon E5-2620 v4 CPUs and two NVIDIA GTX Titan X GPUs with 12 GB memory each.

The evaluation is performed on two public benchmark datasets that are commonly used to evaluate billion-scale ANN search: SIFT1B~\cite{jegou2011searching} of $10^9$ 128-D vectors and DEEP1B~\cite{Yandex2016Efficient} of $10^9$ 96-D vectors. Each dataset has a 10,000 query set with the precomputed ground-truth nearest neighbors. For our system, we sample $2\times10^6$ vectors from each dataset for learning all the trainable parameters. We evaluate the search accuracy by the test result Recall@$K$, which is the rate of queries for which the nearest neighbors is in the top $K$ results.

Here we choose nprobe =64 for all the inverted indexing systems (Faiss, Ivf-hnsw and VLQ-ADC), as 64 is a typical value for nprobe in the Faiss system. The parameter max\_codes that means the max number of candidate data points for a query is only useful to CPU-based system (max\_codes is set to 100,000), hence for GPU-based systems like Faiss and VLQ-ADC, max\_codes parameter is not configured.  In fact, we compute the distances of query point to all the data points that are contained in the neighbor regions.

\subsection{Evaluation without re-ranking}
In experiment 1, we evaluate the index quality of each retrieval system. We compare three different inverted index structures and two inverted multi-index schemes with different codebooks sizes without the re-ranking step.

\begin{enumerate}
  \item
  \textbf{Faiss}. We build a codebook of $k=2^{18}$ centroids by k-means, and find proposed inverted lists of each query by Faiss.
  \item
  \textbf{Ivf-hnsw}. We use a codebook of $k=2^{16}$ centroids  by k-means, and set 64 sub-centroids for each first-level centroid.\footnote{We use the implementation of Ivf-hnsw that is available online \\ (\url{https://github.com/dbaranchuk/ivf-hnsw}) for all the experiments.}
  \item
  \textbf{Multi-D-ADC}. We use two IMI schemes with two codebook sizes $k=2^{10}$ and $k=2^{12}$ and choose the implementation from the Faiss library for all the experiments.
  \item
  \textbf{VLQ-ADC}. For our approach, we use the same codebook as Ivf-hnsw, and a 64-edge k-NN graph with indexing and querying as described in Section \ref{sec:algo} and \ref{sec:query}.\footnote{The VLQ-ADC source code is available at \url{https://github.com/zjuchenwei/vector-line-quantization}.}
\end{enumerate}

The recall curves of each indexing approach are presented in Figure \ref{2p}. On both datasets, our proposed system VLQ-ADC (blue curve) outperforms the other two inverted index systems and the Multi-D-ADC scheme with small codebooks ($k=2^{10}$) for all the reasonable range of $X$. Compared with the Multi-D-ADC scheme with a larger codebook ($k=2^{12}$), our system performs better on DEEP1B, and almost equally well on SIFT1B.

On the DEEP1B dataset, the recall rate of our system is consistently higher than that of all the other indexing structures. With a codebook that is only 1/4 the size of Faiss' codebook, the recall rate of our inverted index is higher than Faiss. This demonstrates that the line quantization procedure can further improve the index quality than the previous inverted index methods.

Even on the SIFT1B dataset, the performance of our indexing structure is almost the same as that of IMI with much larger codebook $k=10^{12}$ and much better than other inverted index structures.

As shown in Figure \ref{2p}, for the SIFT1B dataset, the IMI with $k=2^{12}$ can generate better candidate list than the inverted indexing structures. While for the DEEP1B dataset, the performance of the IMI falls behind that of the inverted indexing structures. The reasons are that SIFT vectors are histogram-based and the subvectors are corresponding to the different subspaces, which describe disjoint image parts that have weak correlations in the subspace distributions. On the other hand, the DEEP vectors are produced by CNN that have a lot of correlations between the subspaces. It can be observed that the performances of our indexing structure is consistent across the two datasets. This demonstrates that our indexing structure's suitability for different data distributions.

\begin{figure*}[!htp]
\centering
    \begin{minipage}[t]{.48\linewidth}
    \centering
    \includegraphics[width=1\textwidth]{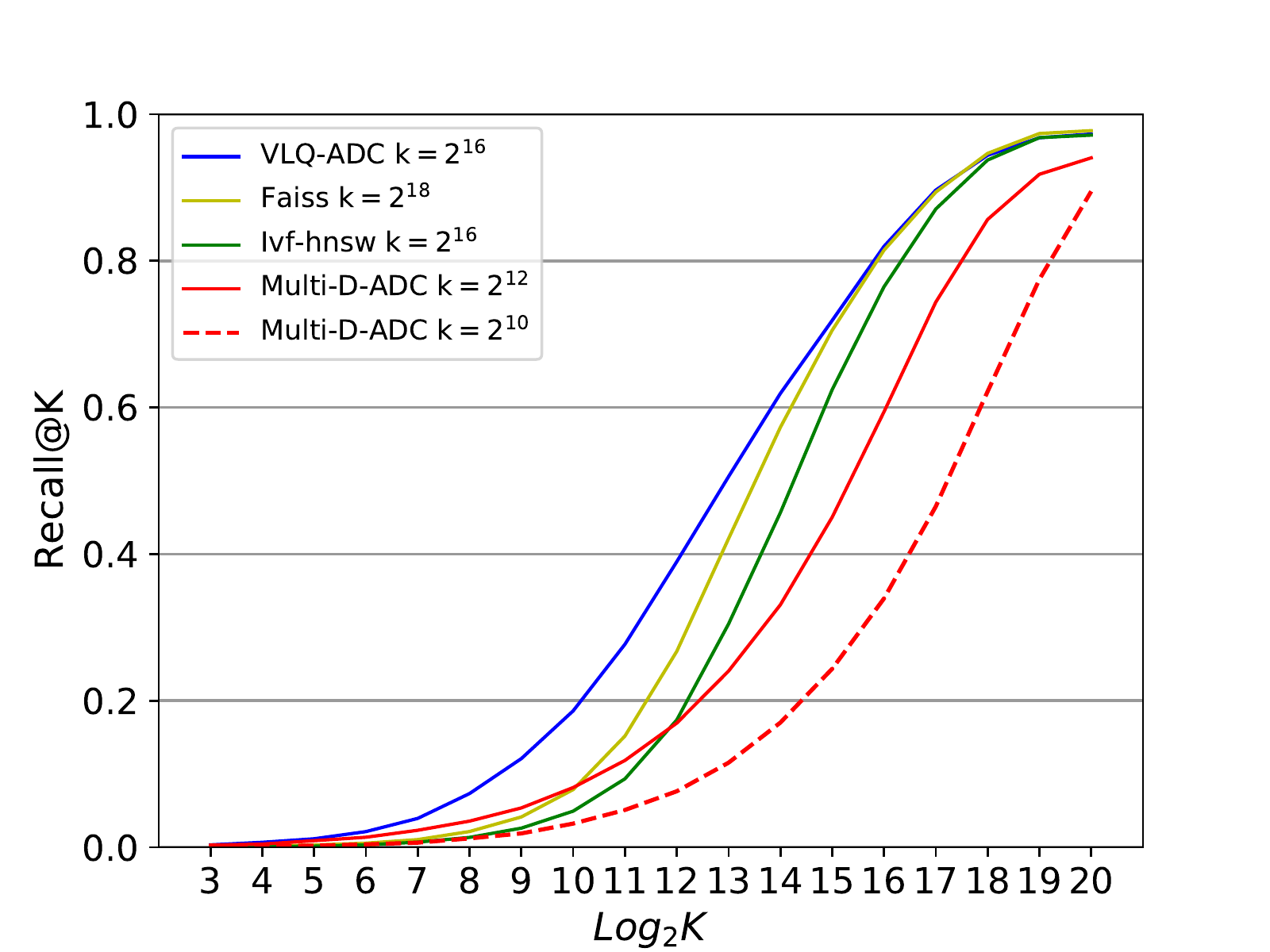}
    \caption*{\footnotesize{DEEP1B}}
    \label{fig:side:deep1b}
    \end{minipage}
    \begin{minipage}[t]{.48\linewidth}
    \centering
    \includegraphics[width=1\textwidth]{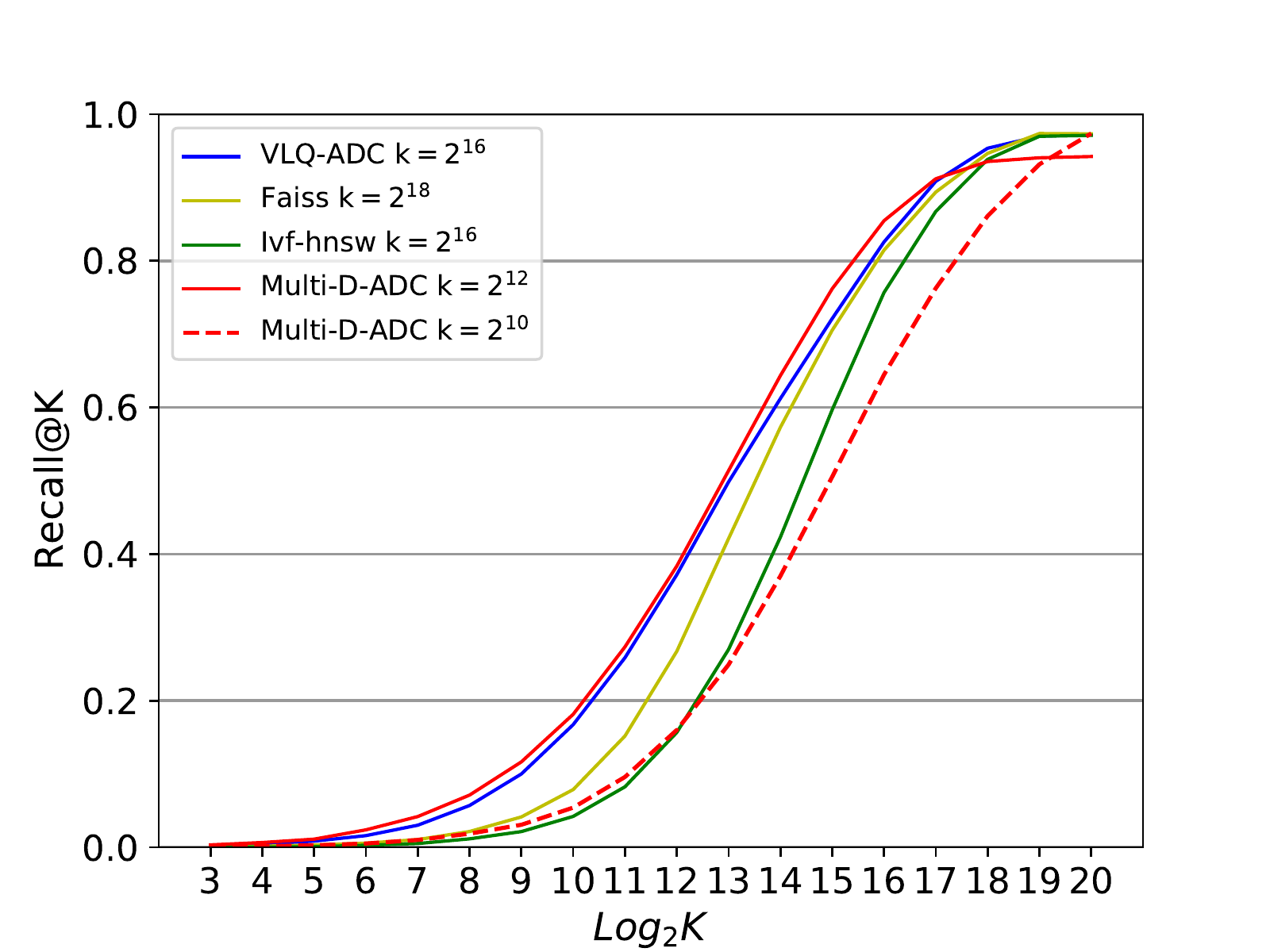}
    \caption*{\footnotesize{SIFT1B}}
    \label{fig:side:sift1b}
    \end{minipage}
    \caption{Recall rate comparison of our system, VLQ-ADC, without the re-ranking step, against two inverted index systems, Faiss, Ivf-hnsw, and one inverted multi-index scheme, Multi-D-ADC (with two different codebook sizes: $k=2^{10}$ and $k=2^{12}$).}
    \label{2p}
\end{figure*}

\subsection{Evaluation with re-ranking}

In experiment 2, we evaluate the recall rates  with the re-ranking step. 
In all systems the dataset points are encoded in the same way: indexing and encoding. (1) \textbf{Indexing}: displacements from data points to the nearest cell centroids are calculated. For VLQ-ADC the displacements are calculated from data points to the nearest anchor points on the line. (2) \textbf{Encoding}: the residual values are encoded into 8 or 16 bytes by PQ with the same codebooks shared by all the cells. Here we compare the same four retrieval systems as in experiments 1. All the configurations for the retrieval systems are the same as in experiment 1. For the GPU-based systems, we evaluate performance with 8-byte codes on 1 GPU and 16-byte codes on 2 GPUs.

The Recall@$K$ values for different values $K=1/10/100$ and the average query times on both datasets in milliseconds (ms) are presented in Table~\ref{table1}. From Table~\ref{table1} we can make the following important observations.

\begin{description}
  \item[Overall best recall performance.]
  Our system VLQ-ADC achieves best recall performance for both datasets and the two codebooks (8-byte and 16-byte) in most cases. For the twelve recall values (Recall@1/10/100 $\times$ two codebooks $\times$ two datasets), VLQ-ADC achieves best values in nine cases and second best in two cases. The second-best system is Faiss, obtaining best results in two cases. Multi-D-ADC (with $k=2^{12}\times2^{12}$ regions) obtains best results in one case.

  \item[Substantial speedup.]
  VLQ-ADC is consistently and significantly faster than all the other systems in all experiments. For all configurations, VLQ-ADC's query time is within 0.054--0.068 milliseconds, while the other systems' query times vary greatly. In the most extreme case, VLQ-ADC is $125\times$ faster than Multi-D-ADC (0.068 vs 8.54). At the same time, VLQ-ADC is also consistently faster than the second fastest system, the GPU-based Faiss, by an average $5\times$ speedup.

  \item[Comparison with Faiss.]
  VLQ-ADC outperforms the current state-of-the-art  GPU-based system Faiss in terms of both accuracy and query time by a large margin, except for only three out of sixteen cases (R@10 with 16-byte codes for SIFT1B, and R@100 with 16-byte codes for SIFT1B and DEEP1B). E.g., as a GPU-based system, VLQ-ADC outperforms Faiss in terms of accuracy by 17\%, 14\%, 4\% of R@1, R@10 and R@100 respectively on the SIFT1B dataset and 8-byte codes. At the same time, the query time is consistently and significantly faster than Faiss, with a speedup of up to $ 5.7 \times $. Faiss outperforms VLQ-ADC in recall values in three cases, all with 16-byte codes. However, the difference is negligible ($\sim$0.02\%). Similarly, though less pronounced,  characteristics can be observed on DEEP1B.

  The main reason for this improvement is that the index quality and encoding precision in VLQ-ADC is better than those of Faiss. Due to the better indexing quality, the inverted list of our system is much shorter than that of Faiss, which results in a much shorter query time.


  Additionally, although the codebook size of our system ($k=2^{16}$) is only $1/4$ of that of Faiss ($k=2^{18}$), our system produces more regions ($2^{22}$) than Faiss ($2^{18}$). Therefore, our system achieves better accuracy as well as memory and runtime efficiency than Faiss.

  \item[Comparison with Multi-D-ADC.]
  The proposed system also outperforms the IMI based system Multi-D-ADC both in terms of accuracy and query time on both datasets. For example, VLQ-ADC leads Multi-D-ADC with codebooks $k=2^{12}$ by 14.2\%, 7.4\%, 1.3\% of R@1, R@10 and R@100 respectively on the SIFT1B dataset and 8-byte codes with up to $ 6.8 \times $ speedup. On the DEEP1B dataset, the advantage of our system is even more pronounced. Similarly, VLQ-ADC outperforms Multi-D-ADC scheme with smaller codebooks $k=2^{10}$ even more significantly, especially in terms of query time, where VLQ-ADC consistently achieves speedups of at least one order of magnitude while obtaining better recall values.

  \item[Comparison with Ivf-hnsw.]
  Similarly, VLQ-ADC outperforms Ivf-hnsw, another CPU-based retrieval system in both recall and query time. Although Ivf-hnsw can also produce more regions with a small codebook, it still cannot outperform the VQ-based indexing structure with larger size of codebook.

  \item[Effects on recall of indexing and encoding.]
  The improvement of R@10 and R@100 shows that the second-level line quantization provides more accurate short-list of candidates than the previous inverted index structure, and the improvement of R@1 shows that it can also improves encoding accuracy.

  \item[Multi-D-ADC.]
  From Table \ref{table1}, we can also observe that Multi-D-ADC scheme with $k=2^{12}$ outperforms the scheme with $k=2^{10}$ in query time by a large margin. It is mainly because Multi-D-ADC with larger codebooks can produce more regions, which can extract more concise and accurate short-lists of candidates.

\end{description}

\begin{table*}[!htb]
\centering
\caption{Performance comparison between VLQ-ADC (with the re-ranking step) against three other state-of-the-art retrieval systems of recall@1/10/100 and retrieval time on two public datasets. For each system the number of total regions is specified beneath each system's name. VLQ-ADC consistently achieves higher recall values and significantly lower query time than all other systems. Best result in each column is \textbf{bolded}, and second best is \underline{underlined}. For the two GPU-based systems, Faiss and VLQ-ADC, we experiments are performed on 1 GPU for 8-byte encoding length, and on 2 GPUs for 16-byte encoding length.}
\label{table1}
\resizebox{\textwidth}{!}{
\setlength\extrarowheight{4pt}
\begin{tabular}{p{78pt}*{16}{r}}
\toprule
\multirow{3}{*}{System} & \multicolumn{8}{c}{SIFT1B} & \multicolumn{8}{c}{DEEP1B} \\
\cmidrule(lr){2-9} \cmidrule(lr){10-17}
 & \multicolumn{4}{c}{8 bytes} & \multicolumn{4}{c}{16 bytes} & \multicolumn{4}{c}{8 bytes} & \multicolumn{4}{c}{16 bytes} \\
\cmidrule(lr){2-5}\cmidrule(lr){6-9}\cmidrule(lr){10-13}\cmidrule(lr){14-17}
 & R@1   & R@10  & R@100  & t (ms)  & R@1   & R@10   & R@100  & t (ms)  & R@1   & R@10  & R@100  & t (ms)  & R@1   & R@10   & R@100  &t (ms)  \\
\midrule
Faiss\newline $2^{18}$ & 0.1383 & 0.4432 & 0.7978 & \underline{0.31} & 0.3180 & 0.7825 & 0.9618 & \underline{0.280} & \underline{0.2101} & \underline{0.4675} & \underline{0.7438} & \underline{0.32} & \underline{0.3793} & \textbf{0.7650} & \textbf{0.9509} & \underline{0.33} \\
Ivf-hnsw\newline $2^{16}$ & \underline{0.1599} & \underline{0.496} & 0.778 & 2.35 & 0.331 & 0.737 & 0.8367 & 2.77 & 0.217 & 0.467 & 0.7195 & 2.30 & 0.3646 & 0.7096 & 0.828 & 3.07 \\
Multi-D-ADC\newline $2^{10} \times 2^{10}$ & 0.1255 & 0.4191 & 0.7843 & 1.65 & 0.3064 & 0.7716 & \textbf{0.9782} & 8.54 & 0.1716 & 0.3834 & 0.6527 & 3.28 & 0.324 & 0.6918 & 0.9258 & 6.152 \\
Multi-D-ADC\newline $2^{12} \times 2^{12}$ & 0.1420 & 0.4720 & \underline{0.8183} & 0.367 & \underline{0.3324} & \underline{0.8029} & \underline{0.9752} & 1.603 & 0.1874 & 0.4240 & 0.6979 & 0.839 & 0.3557 & 0.7087 & 0.9059 & 1.52 \\
\midrule
VLQ-ADC\newline $2^{16}$ & \textbf{0.1620} & \textbf{0.507} & \textbf{0.829} & \textbf{0.054} & \textbf{0.345} & \textbf{0.8033} & 0.9400 & \textbf{0.068} & \textbf{0.2227} & \textbf{0.4855} & \textbf{0.7559} & \textbf{0.059} & \textbf{0.394} & \underline{0.7644} & \underline{0.9272} & \textbf{0.067}\\
\bottomrule
\end{tabular}
}
\end{table*}

\begin{figure*}[!htb]
\centering
    \begin{minipage}[t]{0.46\linewidth}
    \centering
    \includegraphics[width=\textwidth]{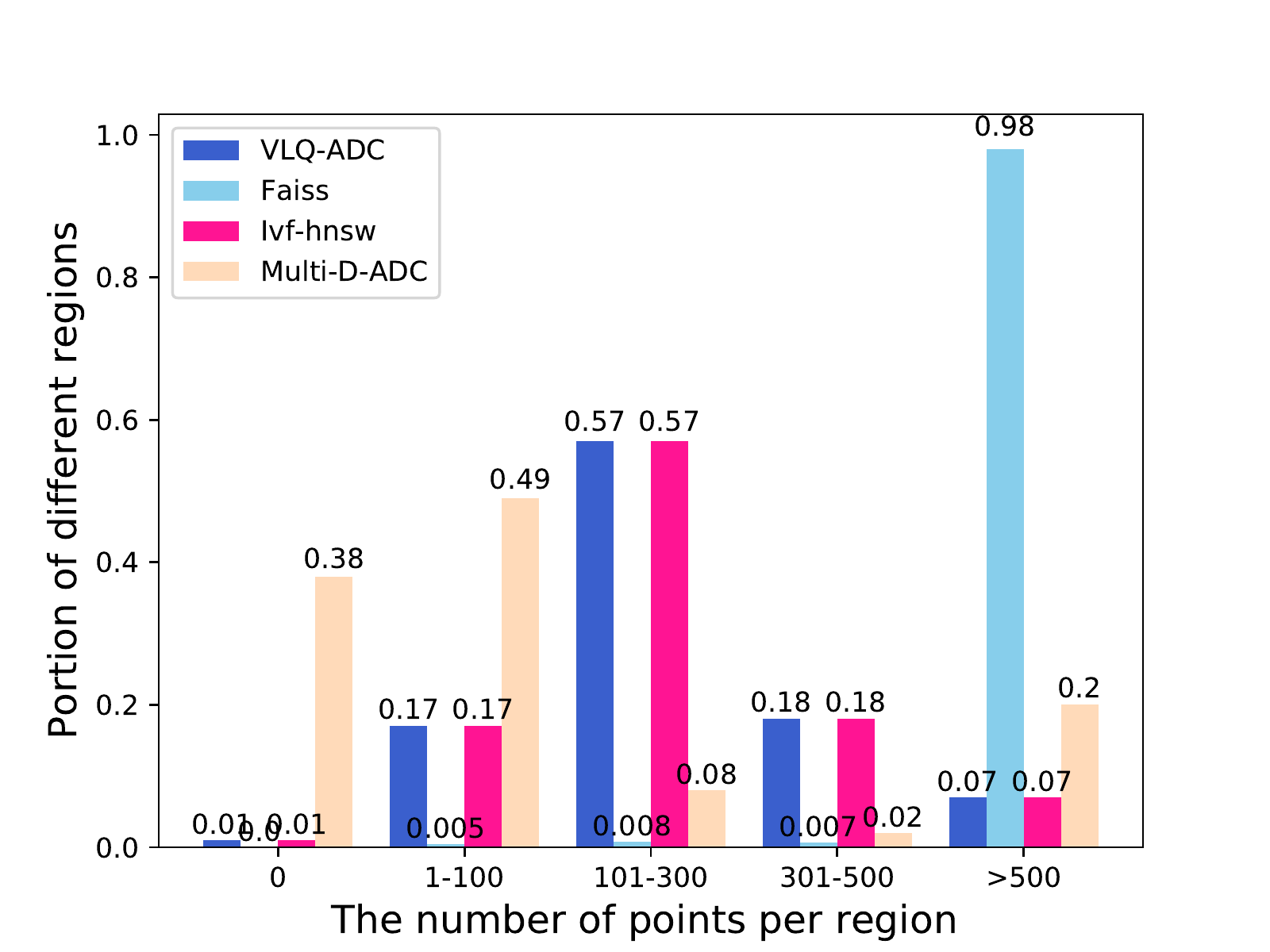}
    \caption*{\footnotesize{SIFT1B}}
    \label{fig:side:a}
    \end{minipage}
    \begin{minipage}[t]{0.46\linewidth}
    \centering
    \includegraphics[width=\textwidth]{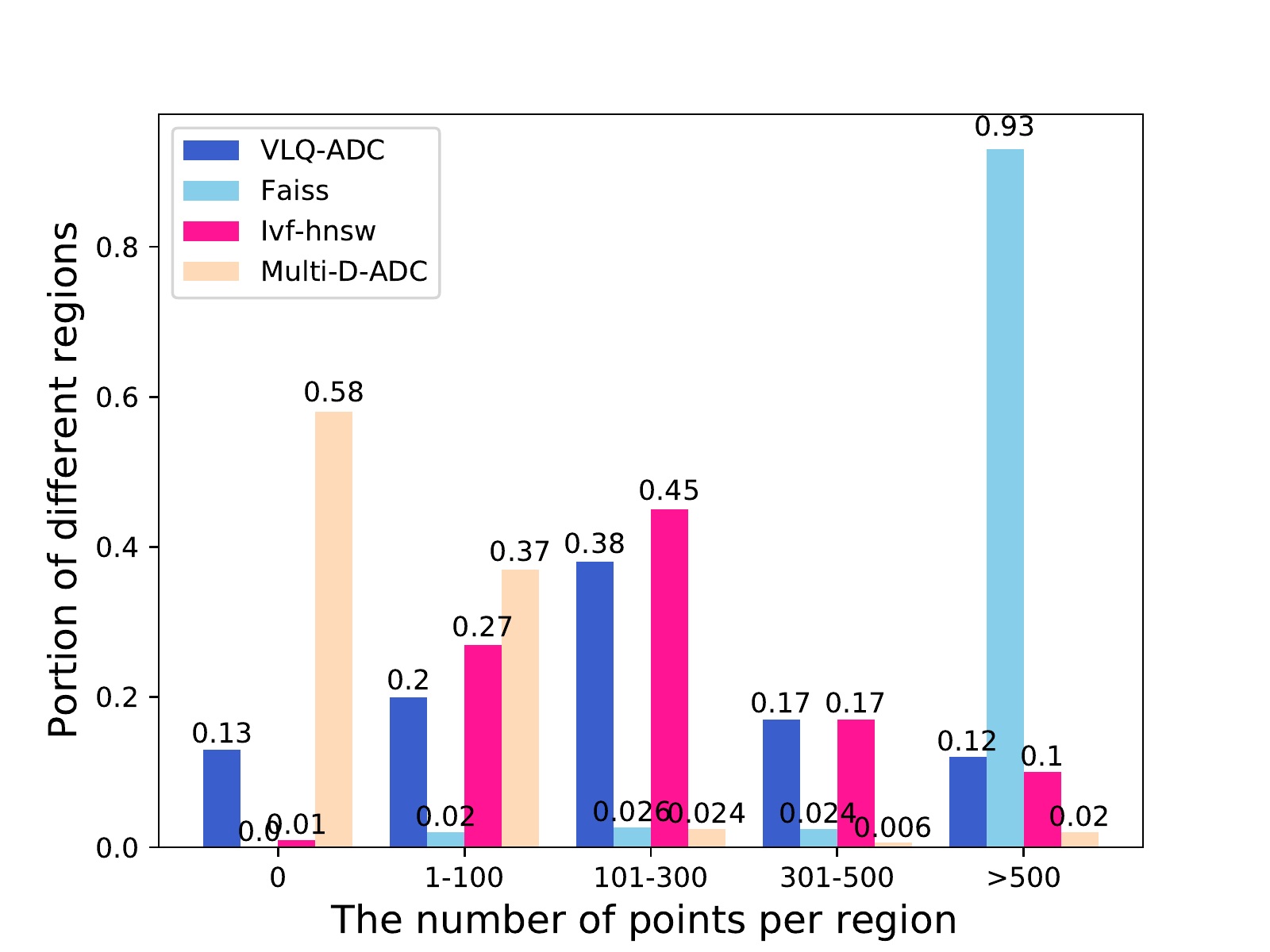}
    \caption*{\footnotesize{DEEP1B}}
    \label{fig:side:b}
    \end{minipage}
    \caption{The distributions of data points in regions produced by different indexing structures. The x axis is five categories representing the discretized numbers of data points in each region (0, 1--100, 101--300, 301--500 and $>500$). The y axis is the percentage of regions in each different categories. }
    \label{fig:d_values}
\end{figure*}

\subsection{Data point distributions of different indexing structures}
The space and time efficiency of an indexing structure is impacted by the distribution of data points produced by the structure. To analyse the distribution produced by the structures studied in this paper, we plot in Figure~\ref{fig:d_values} the percentages of regions by the discretized number of data points in each region.

As shown in Figure \ref{fig:d_values}, the portion of empty regions produced by the inverted indexing structures (Faiss, Ivf-hnsw and VLQ-ADC) is much less than that produced by the inverted multi-index structure (Multi-D-ADC). For Multi-D-ADC, there are 38\% empty regions for SIFT1B and 58\% empty regions for DEEP1B (left most group in each plot). This result empirically validates the space inefficiency of inverted multi-index structure~\cite{babenko2012inverted}.

For Faiss, which is based on the inverted indexing structure using VQ, over 98\% and 93\% of regions contain more than 500 data points for SIFT1B and DEEP1B respectively. This will possibly produce long candidate lists for queries, thus negatively impacting query speed. For VLQ-ADC (and Ivf-hnsw), the regions are much more evenly distributed. The majority of the regions on both datasets contain less than 500 data points, and more regions contain 101--300 data points than others. This is a main reason why VLQ-ADC can provide shorter candidate lists and thus a faster query speed.

\begin{figure*}[!htb]
\centering
    \begin{minipage}[t]{0.49\linewidth}
    \centering
    \includegraphics[width=\textwidth]{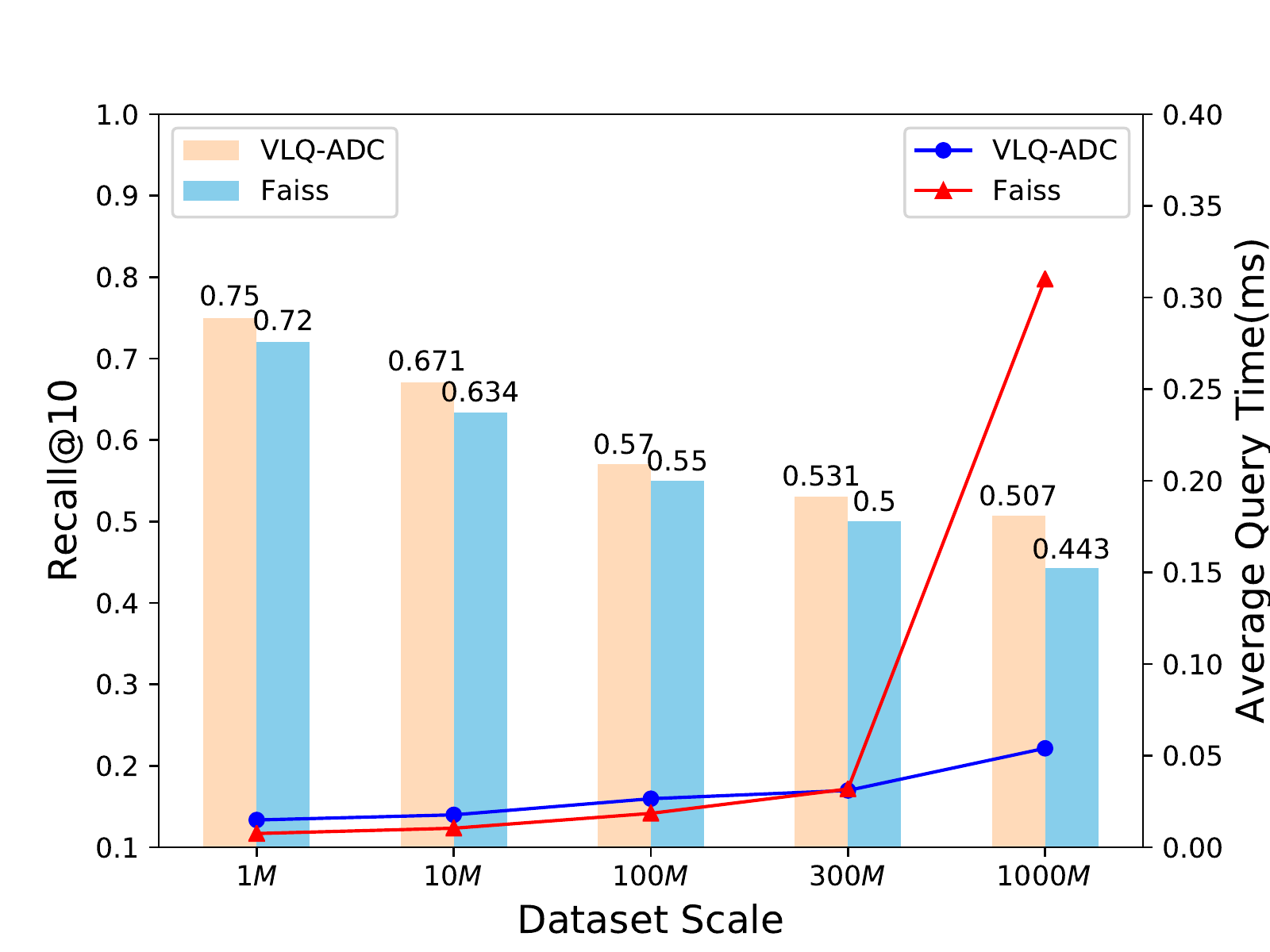}
    \caption*{\footnotesize{SIFT1B}}
    \label{fig:k:a}
    \end{minipage}
    \begin{minipage}[t]{0.49\linewidth}
    \centering
    \includegraphics[width=\textwidth]{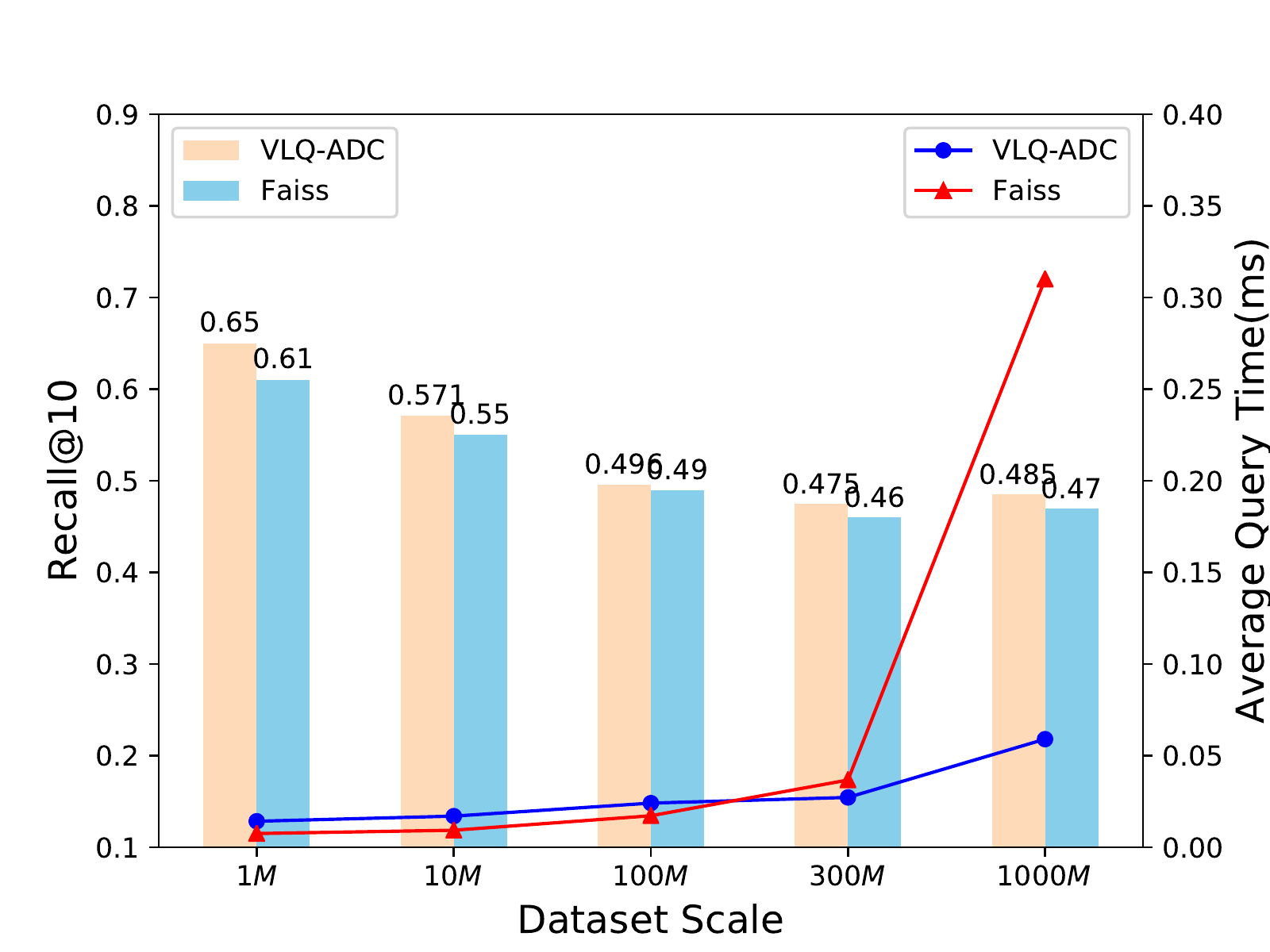}
    \caption*{\footnotesize{DEEP1B}}
    \label{fig:k:b}
    \end{minipage}
    \vspace{-6pt}
    \caption{Comparison of recall@10 and average query time between \sys and Faiss under different dataset scales. The two systems are compared with an 8-byte encoding length. The $x$ axis indicates the five data scales (1M/10M/100M/300M/1000M). The left $y$ axis is the recall@10 value (represented by the bars) and the right $y$ axis is the average query time (in ms, represented by the lines).}
    \label{fig:k1_values}
    \vspace{-6pt}
\end{figure*}

\vspace{-6pt}
\subsection{Performance comparison under different dataset scales}
 In this section we evaluate the performance of our system under different dataset scales. Figure \ref{fig:k1_values} shows, for SIFT1B and DEEP1B, the recall and query time values for Faiss and \sys for subsets of SIFT and DEEP1B of different sizes: 1M, 10M, 100M, 300M and 1000M (full dataset) respectively. As can be seen in the figure, the recall values of \sys is always higher than that of Faiss under all dataset scales. When the scale of dataset is under 300M, the query speed of Faiss is slightly faster than that of \sys. When the scale of the datasets is over 300M, the query speed of \sys matches that of Faiss.

 It can also been observed from the figure that for the full datasets of SIFT1B and DEEP1B (1000M), Faiss takes 0.31ms and 0.32ms respectively (see Table~\ref{table1} too). Compared to the 100M subsets of these two datasets, Faiss suffers an approx.\ $15\times$ slowdown when data scale grows $10\times$. On the other hand, for these two datasets, \sys takes 0.054ms and 0.059ms respectively, representing only an approx.\ $2\times$ slowdown when data scale grows $10\times$. The superior scalability and robustness of \sys over Faiss is evident from this experiment.

\begin{figure*}[!htb]
\centering
    \begin{minipage}[t]{0.32\linewidth}
    \centering
    \includegraphics[width=\textwidth]{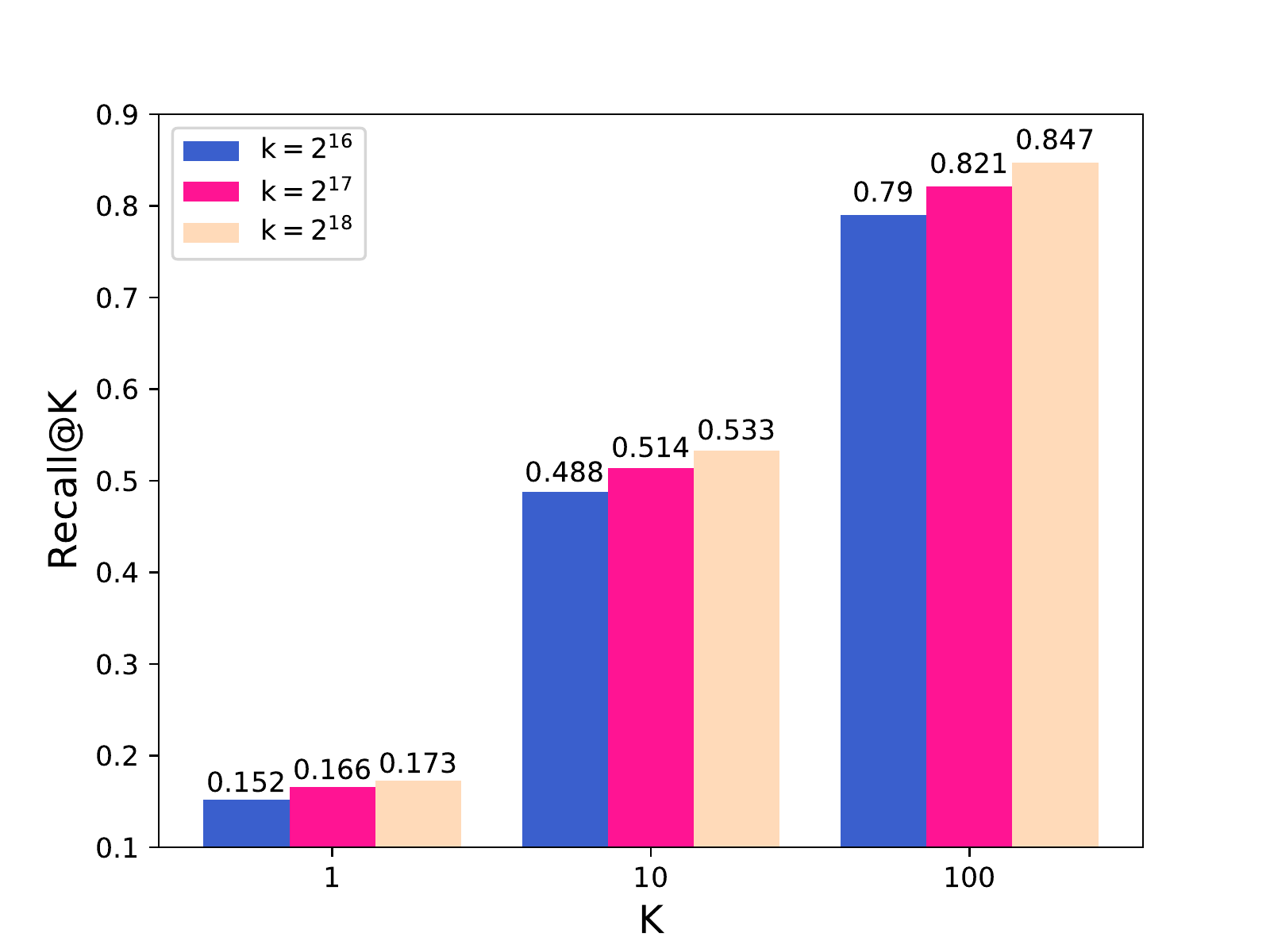}
    \caption*{\footnotesize{SIFT1B}}
    \label{fig:side:a}
    \end{minipage}
    \begin{minipage}[t]{0.32\linewidth}
    \centering
    \includegraphics[width=\textwidth]{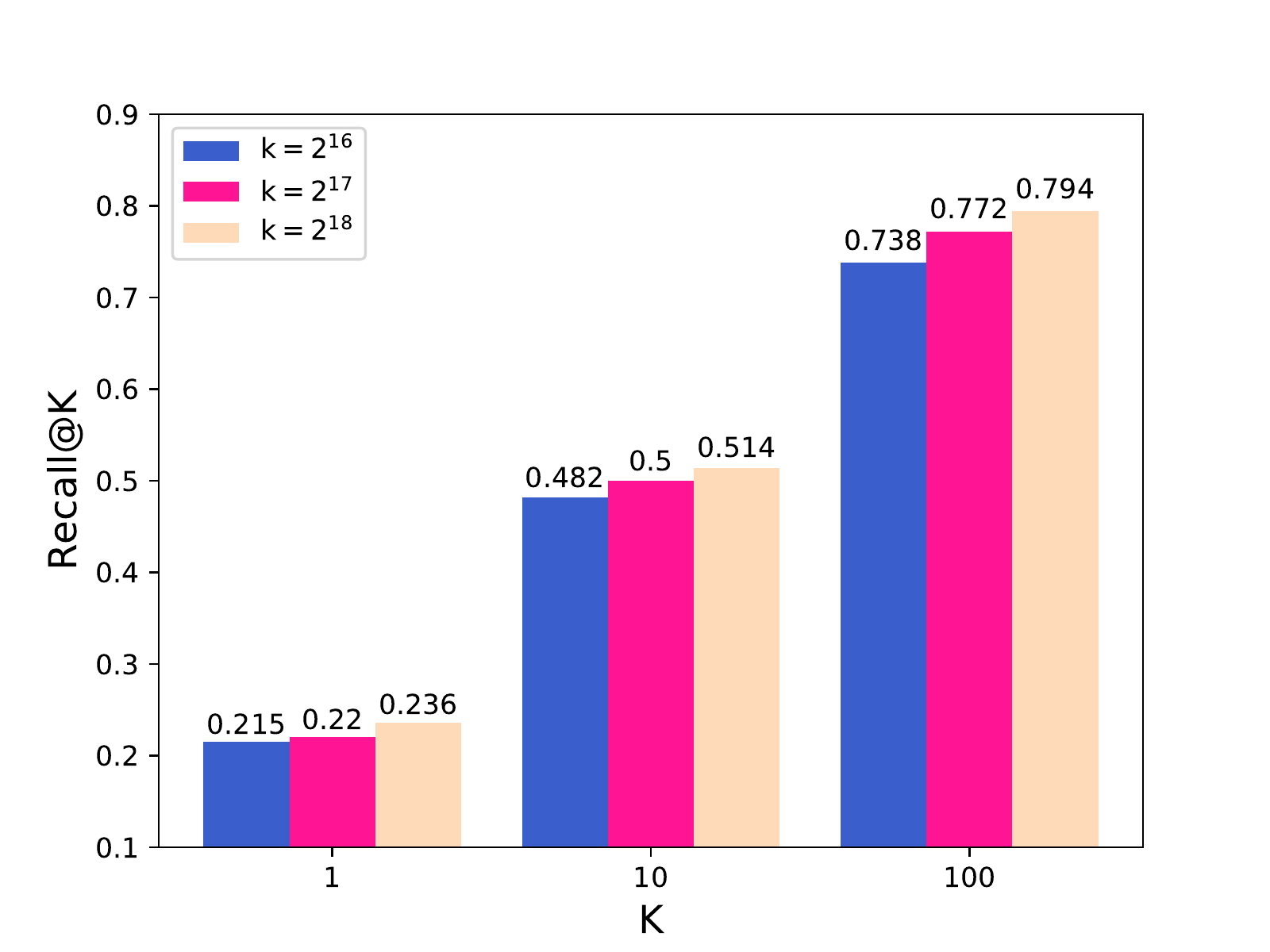}
    \caption*{\footnotesize{DEEP1B}}
    \label{fig:side:b}
    \end{minipage}
    \begin{minipage}[t]{0.32\linewidth}
    \centering
    \includegraphics[width=\textwidth]{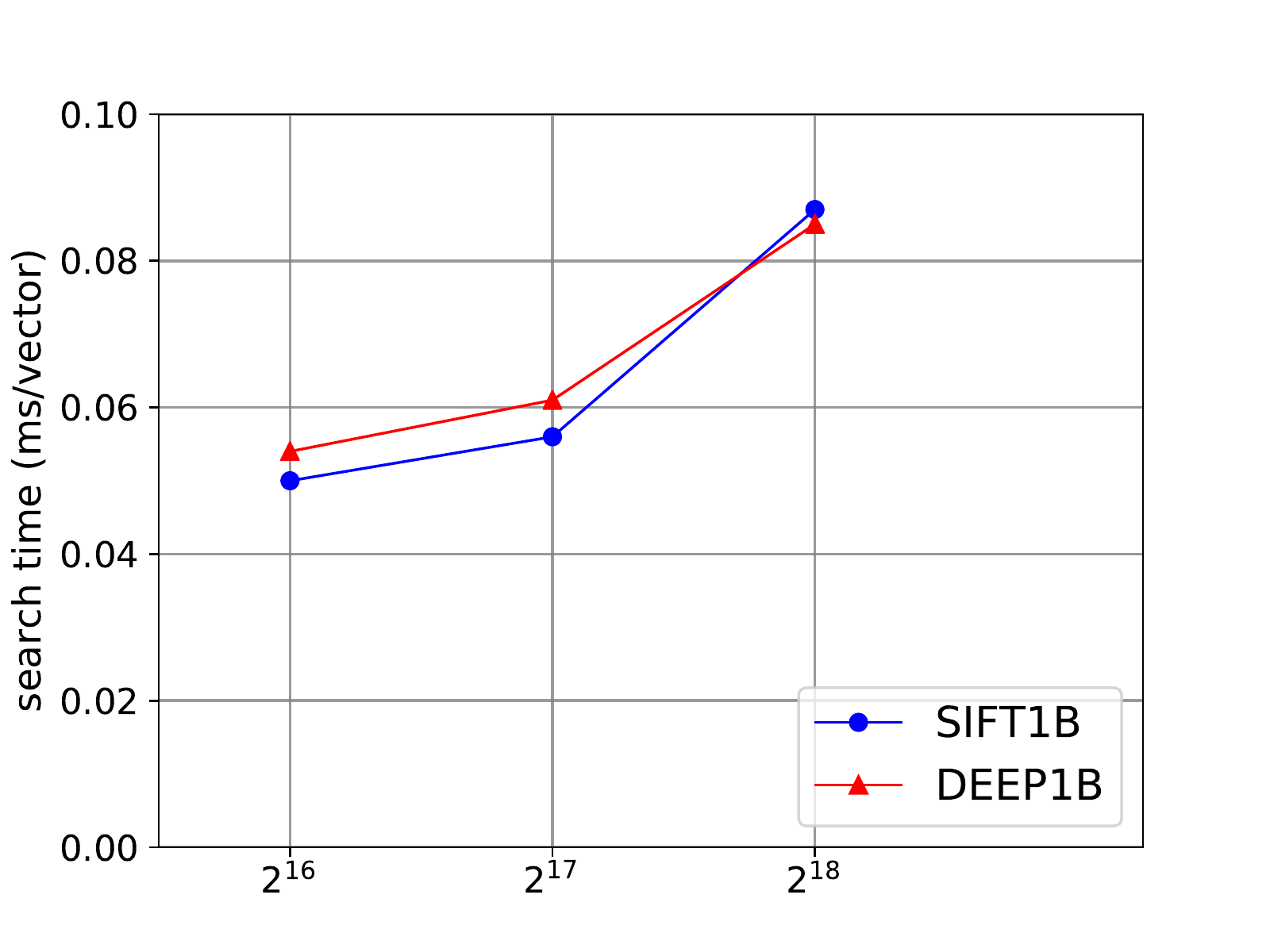}
    \caption*{\footnotesize{Search time}}
    \label{fig:side:c}
    \end{minipage}
    \caption{The performance of VLQ-ADC on different numbers of centroids $k=2^{16}/2^{17}/2^{18}$. The results are collected on the same two datasets with an 8-byte encoding length and $n = 32$ edges of each centroids. The right plot shows the average search time with different values of $k$. }
    \label{fig:k_values}
\end{figure*}

\begin{figure*}[!htb]
\centering
    \begin{minipage}[t]{0.32\linewidth}
    \centering
    \includegraphics[width=\textwidth]{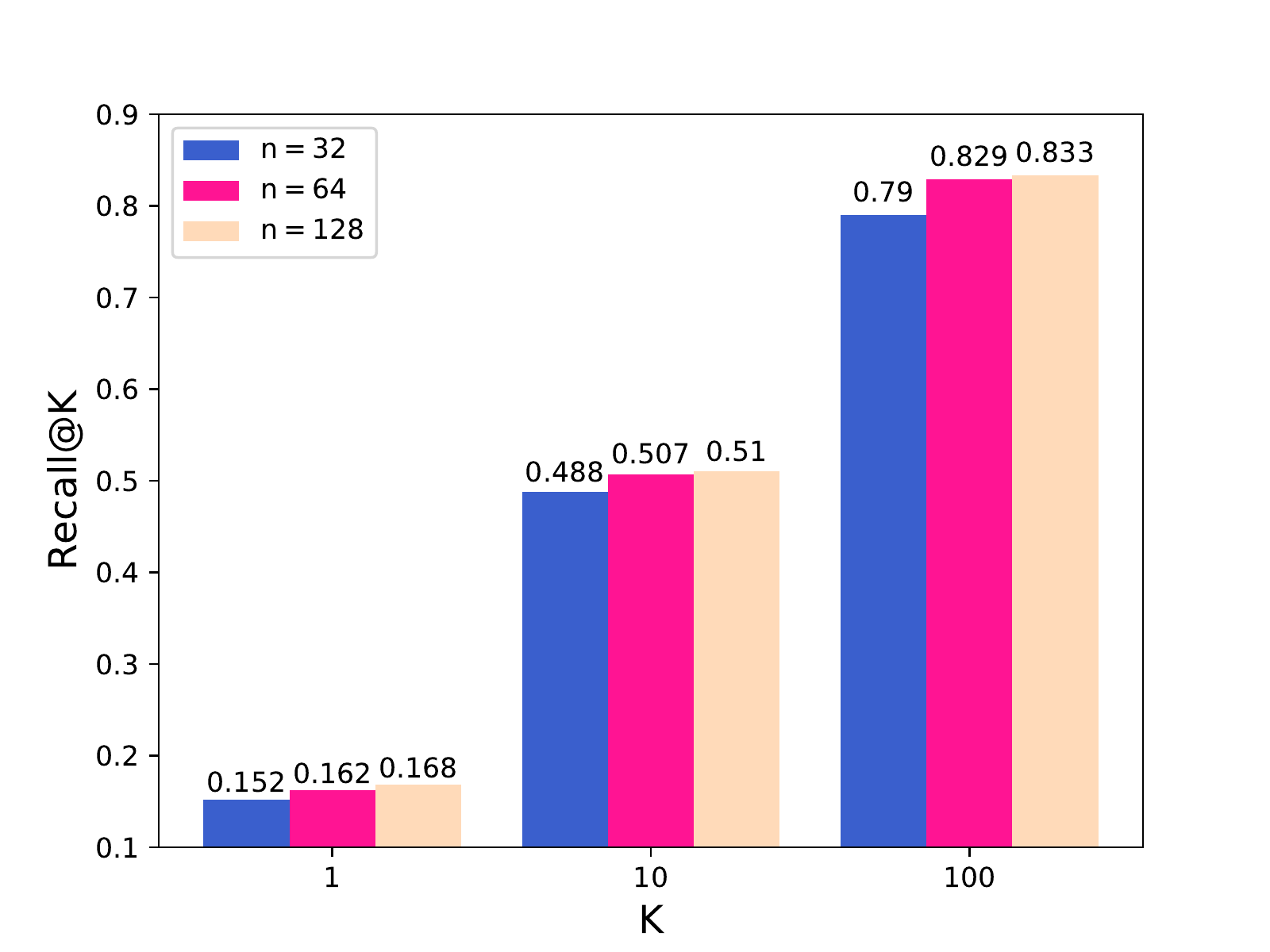}
    \caption*{\footnotesize{SIFT1B}}
    \label{fig:side:a}
    \end{minipage}
    \begin{minipage}[t]{0.32\linewidth}
    \centering
    \includegraphics[width=\textwidth]{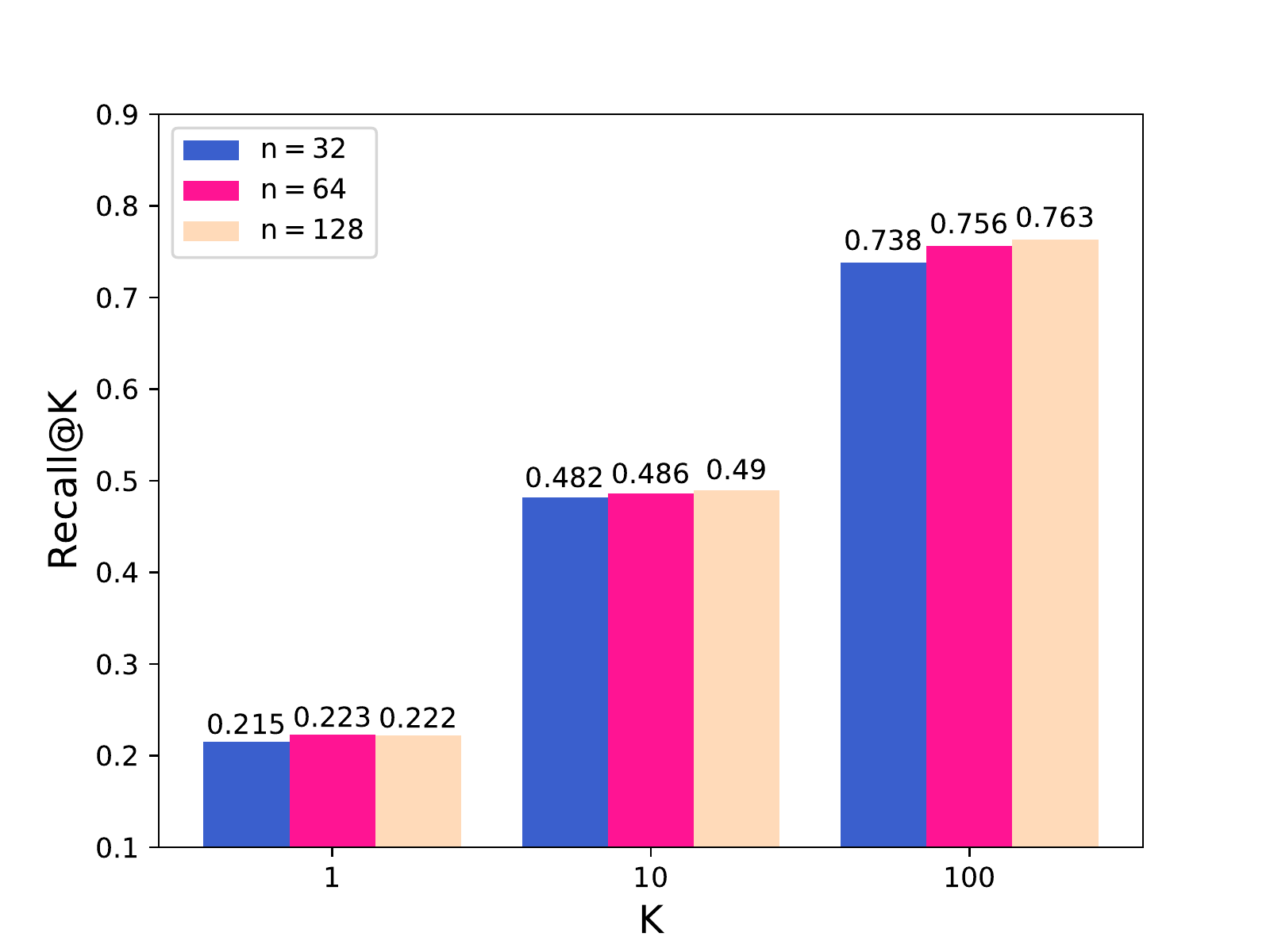}
    \caption*{\footnotesize{DEEP1B}}
    \label{fig:side:b}
    \end{minipage}
    \begin{minipage}[t]{0.32\linewidth}
    \centering
    \includegraphics[width=\textwidth]{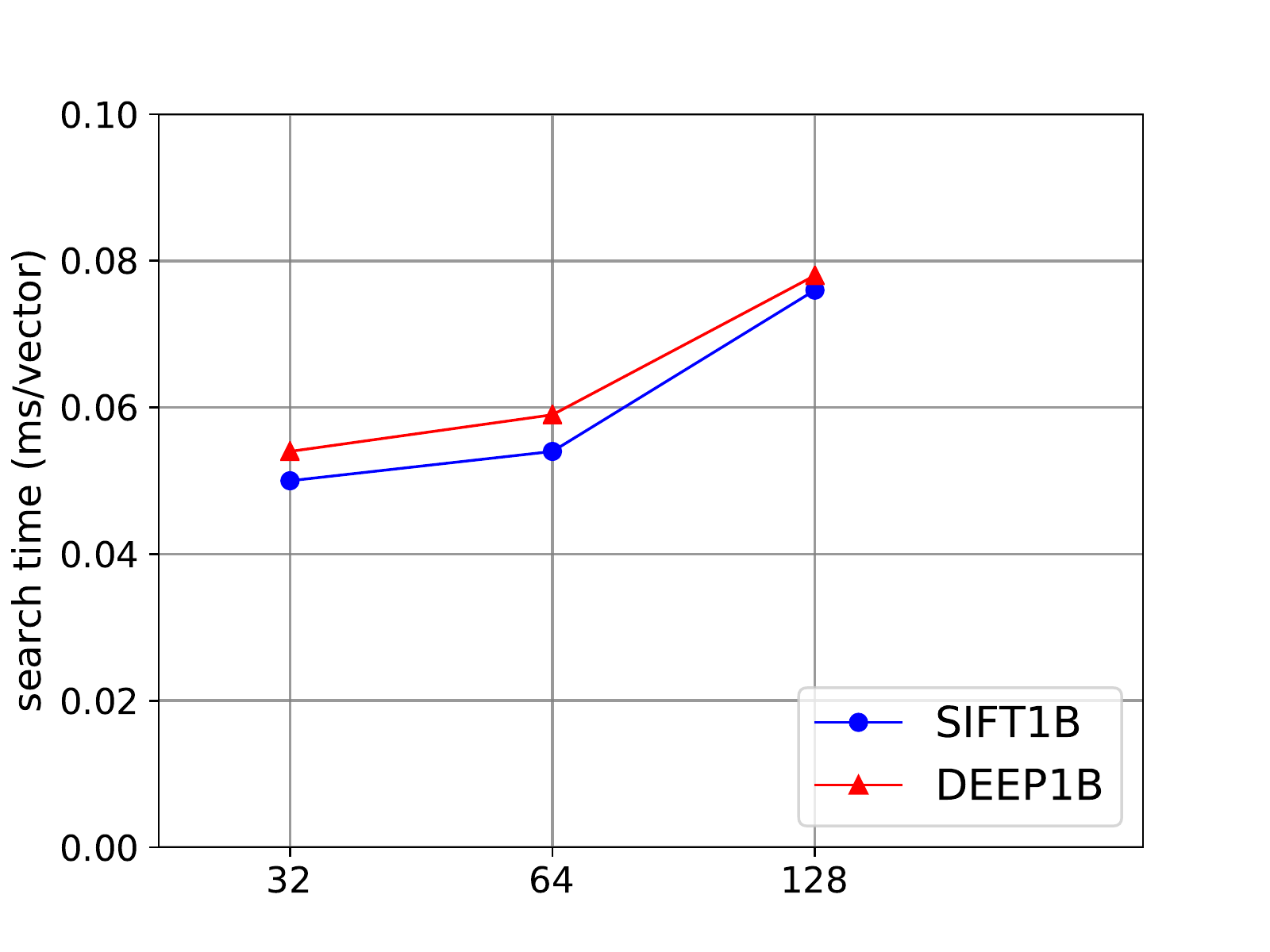}
    \caption*{\footnotesize{Search time}}
    \label{fig:side:c}
    \end{minipage}
    \caption{The performance of VLQ-ADC on different numbers of graph edges $n=32/64/128$. The results are collected on the same two datasets with an 8-byte encoding length and $k = 2^{16}$ number of centroids. The right plot shows the average search time with different values of $n$.}
    \label{fig:n_values}
\end{figure*}

\begin{figure*}[!htb]
\centering
    \begin{minipage}[t]{0.32\linewidth}
    \centering
    \includegraphics[width=\textwidth]{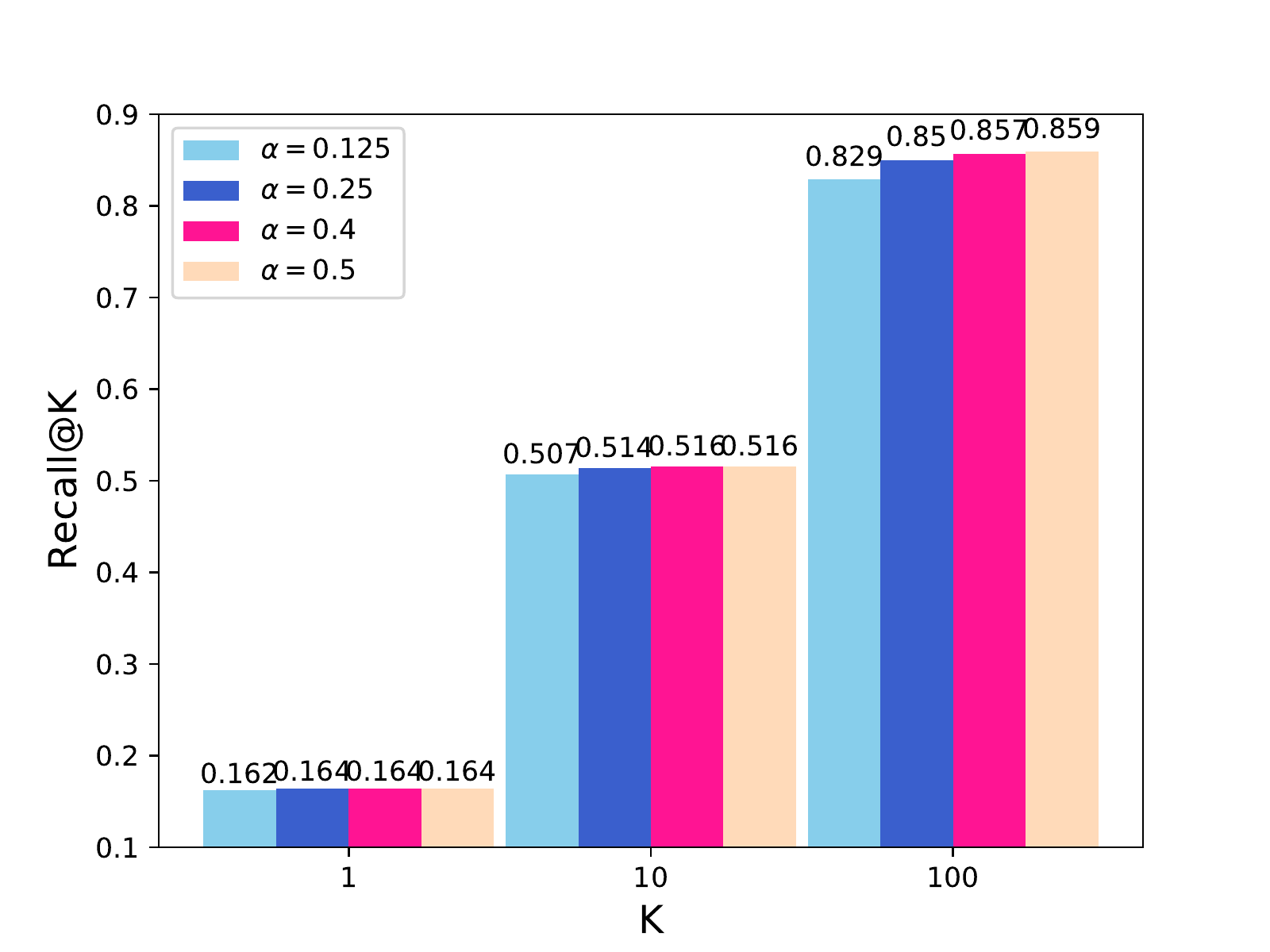}
    \caption*{\footnotesize{SIFT1B}}
    \label{fig:side:a}
    \end{minipage}
    \begin{minipage}[t]{0.32\linewidth}
    \centering
    \includegraphics[width=\textwidth]{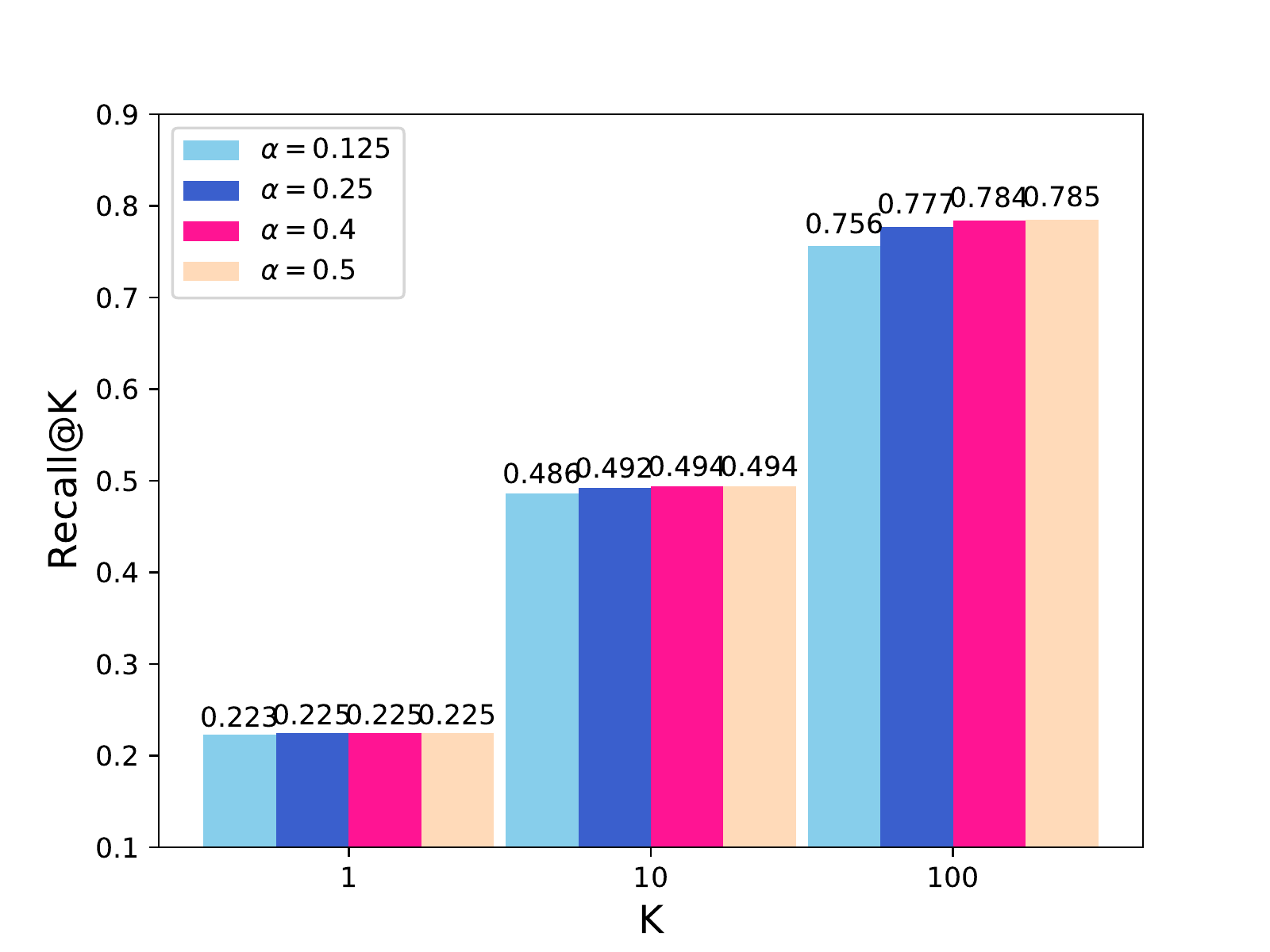}
    \caption*{\footnotesize{DEEP1B}}
    \label{fig:side:b}
    \end{minipage}
    \begin{minipage}[t]{0.32\linewidth}
    \centering
    \includegraphics[width=\textwidth]{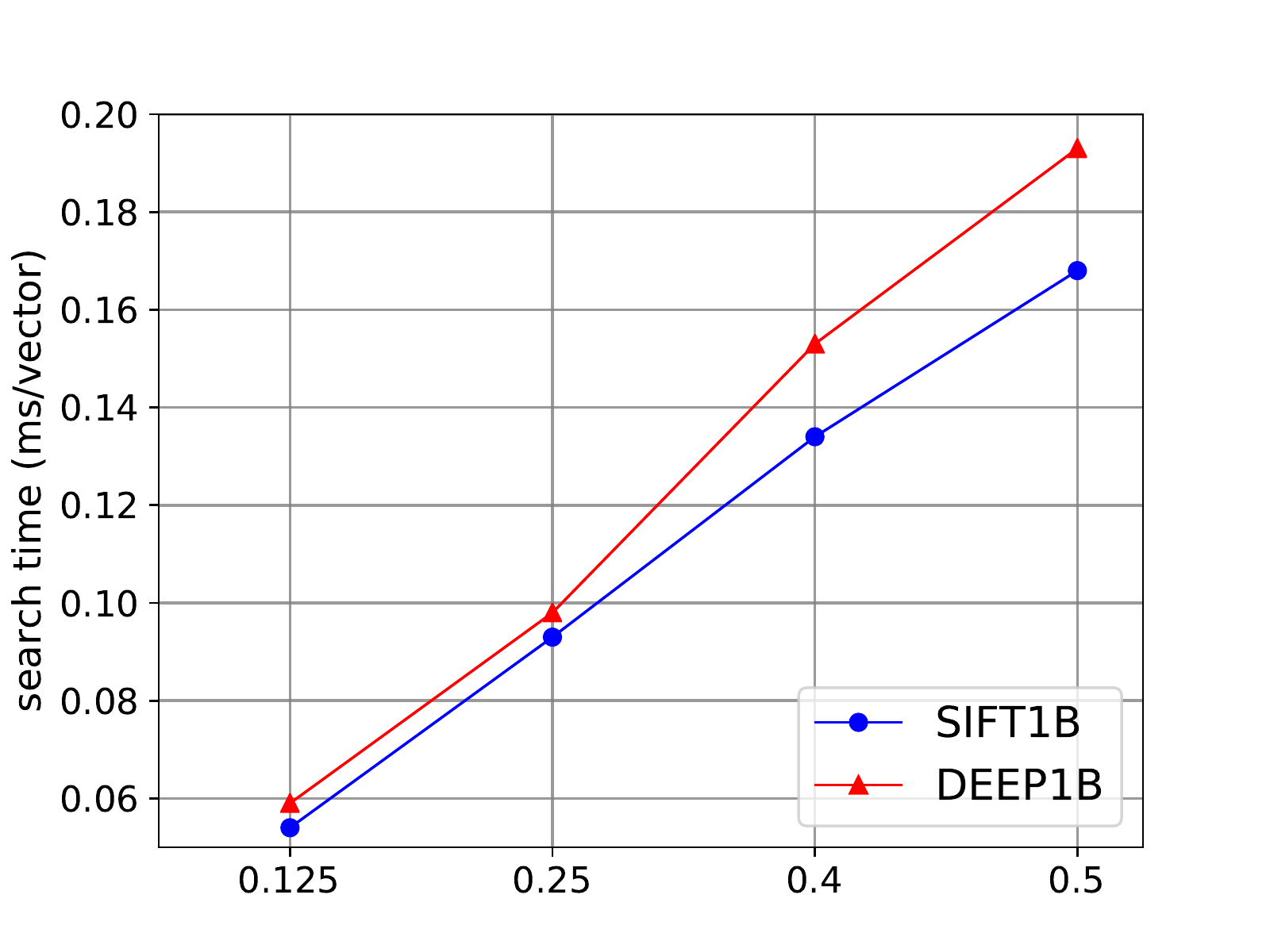}
    \caption*{\footnotesize{Search time}}
    \label{fig:side:c}
    \end{minipage}
    \caption{The performance of VLQ-ADC on different values of parameter $\alpha=0.25/0.4/0.5$, with values of $k$, $n$ and $w$ fixed at $k=2^{16},n=64,w=64$. The result are collected on the same two datasets with an 8-byte encoding length and 64 edges of each centroids. The right plot shows the average search time with different values of $\alpha$. }
    \label{fig:a_values}
\end{figure*}

\subsection{Evaluation on impact of parameter values}

\noindent\textbf{Number of centroids $k$ and  edges $n$.} We evaluate the performance of VLQ-ADC on different $k$ and $n$ values with 8-byte codes. We first fix the value of $n$ to 64 and compare the performance of our system for different $k$ centroids. In Figure~\ref{fig:k_values}, we present the evaluation of VLQ-ADC for $k=2^{16}/2^{17}/2^{18}$. Then we fix $k=2^{16}$ and increase the number of edge $n$ from 32 to 64 and 128. In Figure~\ref{fig:n_values}, we present the evaluation of the VLQ-ADC for different edge numbers.

From Figure~\ref{fig:k_values} and~\ref{fig:n_values} we can observe that the increase in the number of centroids and edges can improve search accuracy, while slightly increasing query time. This is because the indexing scheme with more centroids and more edges can represent the dataset points more accurately and hence provide more accurate short inverted lists.

\noindent\textbf{Value of portion $\alpha$.} Now, we discuss how to determine the value of parameter $\alpha$ for subregions pruning, as described in Section~\ref{s3.3.1}. As shown in Figure \ref{fig:a_values}, we test several values of $\alpha$ on both datasets. A lower $\alpha$ value means fewer subregions will be traversed, hence lower query time. At the same time, we can observe that higher $\alpha$ values only moderately increase recall values, while significantly increases query time (up to $3.7 \times$ times). Hence we choose $\alpha=0.25$.

\noindent\textbf{Time and memory consumption.} Because the billion-scale dataset  do not fit on the GPU, the database is built in batches of 2M vectors,  then aggregating the information on the CPU. With file I/O, it takes about 150 minutes to build the whole database on a single GPU.

Here we analyze the memory consumption of each system. As shown in Table \ref{tab:sum}, for a database of $N=10^9$ points,  the basic memory consumption for all systems is $4\cdot N$ bytes for point IDs that are Integer type and $m\cdot N$ bytes for point codes.
In addition to that, Multi-D-ADC consumes $4\cdot k^2$ bytes to store the region boundaries. Faiss consumes $4\cdot k \cdot D$ bytes for the codebooks and $4\cdot k \cdot m  \cdot 256$ bytes for the lookup tables. Ivf-hnsw requires $N$ bytes for quantized norm items $4\cdot k \cdot (D+n)$ bytes for its indexing structure\cite{Baranchuk2018Revisiting}. For our system, we require $N$ bytes for quantized $\lambda$ values and $4\cdot k \cdot (D+2n+m \cdot 256)$ bytes for the codebook, the $n$-NN graph and the lookup tables. We summarize the total memory consumption for all systems in Table \ref{tab:m_values} with 8-byte encoding length on both datasets.

As presented in Table \ref{tab:m_values}, the memory consumption of our system is less than that of Faiss, and about 10\% more than that of Multi-D-ADC with $2^{12}$ codebook, which is acceptable for most realistic setups.

\begin{table}[htb]
\centering
\caption{The memory consumption of all systems for SIFT1B of $10^9$ 128-dimensional data points.}
\label{tab:m_values}
\begin{tabular}{lr}
\toprule
System (codebook size) & Memory consumption (GB) \\
\midrule
Faiss ($2^{18}$)  &  14       \\
Ivf-hnsw ($2^{16}$) &  13.04      \\
Multi-D-ADC ($2^{12} \times 2^{12}$) & 12.25     \\
\midrule
VLQ-ADC  ($2^{16}$) &  13.55   \\
\bottomrule
\end{tabular}
\end{table}

\section{Conclusion}
\label{s6}
Billion-scale approximate nearest neighbor (ANN) search has become an important task as massive amounts of visual data becomes available online. In this work, we proposed VLQ-ADC, a simple yet scalable indexing structure and a retrieval system that is capable of handling billion-scale datasets. VLQ-ADC combines line quantization with vector quantization to create a hierarchical indexing structure. Search space is further pruned to a portion of the closest regions, further improving ANN search performance. The proposed indexing structure can partition the billion-scale database in large number of regions with a moderate size of codebook, which solved the drawback of prior VQ-based indexing structures.

We performed comprehensive evaluation on two billion-scale benchmark datasets: SIFT1B and DEEP1B and three state-of-the-art ANN search systems: Multi-D-ADC, Ivf-hnsw, and Faiss. Our evaluation shows that VLQ-ADC consistently outperforms all three systems on both recall and query time. VLQ-ADC achieves a recall improvement over Faiss, the state-of-the-art GPU-based system, of up to 17\% and a query time speedup of up to $ 5 \times $ times.

Moreover, VLQ-ADC takes the data distribution into account in the indexing structure. As a result, it performs well on datasets with different distributions. Our evaluation shows that VLQ-ADC is the best performer on both SIFT1B and DEEP1B, demonstrating its robustness with respect to data with different distributions.

We conclude by pointing out a number of future work directions. We plan to investigate further improvements to the indexing structure. Moreover, a more systematic and principled method for hyperparameter selection is worthy investigation.

\section*{Acknowledgment}
This work is supported in part by the National Natural Science Foundation of China under Grant No.61672246, No.61272068, No.61672254 and the Fundamental Research Funds for the Central Universities under Grant
HUST:2016YXMS018. In addition, we gratefully acknowledge the support of NVIDIA Corporation with the donation of the Titan Xp GPUs used for this research. The authors appreciate the valuable suggestions from the anonymous reviewers and the Editors.




\end{document}